\documentclass[conference]{IEEEtran}
\IEEEoverridecommandlockouts
\usepackage{times}

\usepackage[numbers]{natbib} 
\usepackage{multicol} 
\usepackage[bookmarks=true,pdfencoding=auto,breaklinks,colorlinks,hypertexnames=false]{hyperref} 
\usepackage{graphicx}
\usepackage{amsmath}
\usepackage{amssymb}
\usepackage{booktabs}
\usepackage{xcolor}
\usepackage[flushleft]{threeparttable}
\usepackage{multirow, bigdelim}
\usepackage{url}

\usepackage{siunitx}
\usepackage{flushend}
\usepackage{colortbl}
\usepackage{pifont}
\usepackage{cuted}
\usepackage{tcolorbox}

\usepackage{caption}
\usepackage{subcaption}
\usepackage{pgfplots}
\pgfplotsset{width=7cm, compat=1.10}
\usepgfplotslibrary{fillbetween}

\begin{document}

\title{Hierarchical Open-Vocabulary 3D Scene Graphs \\ for Language-Grounded Robot Navigation}


\author{\authorblockN{
Abdelrhman Werby\textsuperscript{\footnotesize 1*},\quad
Chenguang Huang\textsuperscript{\footnotesize 1*},\quad
Martin Büchner\textsuperscript{\footnotesize 1*},\quad
Abhinav Valada\textsuperscript{\footnotesize 1}, \quad
Wolfram Burgard\textsuperscript{\footnotesize 2}}
\vspace{0.3cm}
\authorblockA{\textsuperscript{\footnotesize 1}University of Freiburg \quad\quad {\textsuperscript{\footnotesize 2}University of Technology Nuremberg}}
\thanks{$^{*}$ Equal contribution.}%
} 


\makeatletter
\newcommand\blfootnote[1]{%
  \begingroup
  \renewcommand\thefootnote{}\footnote{#1}%
  \addtocounter{footnote}{-1}%
  \endgroup
}
\makeatother

\newcommand{\red}[1]{\textcolor{red}{#1}}

\newcommand{\refeqn}[1]{Eq.~\ref{#1}}
\newcommand{\refEqn}[1]{Equation~\ref{#1}}
\newcommand{\reffig}[1]{Fig.~\ref{#1}}
\newcommand{\refsec}[1]{Sec.~\ref{#1}}
\newcommand{\refalg}[1]{Algorithm~\ref{#1}}
\newcommand{\reftab}[1]{Table~\ref{#1}}

\newcommand{\greyrule}{\arrayrulecolor{black!30}\midrule\arrayrulecolor{black}}

\newcommand{\ours}{HOV-SG}
\newcommand{\website}{\url{https://hovsg.github.io}}

\newcommand{\chenguang}[1]{\textcolor{violet}{@chenguang #1}}
\newcommand{\martin}[1]{\textcolor{purple}{@martin #1}}
\newcommand{\werby}[1]{\textcolor{teal}{@werby #1}}

\newcommand{\rebuttal}[1]{\textcolor{black}{#1}}

\newcommand{\xmark}{\ding{55}}%
\newcommand{\cmark}{\ding{51}}  
\newcommand{\cross}{\ding{61}}

\definecolor{light-gray}{rgb}{0.8, 0.8, 0.8}
\definecolor{highlight}{HTML}{e3eeff}
\definecolor{comment-green}{rgb}{0.435, 0.576, 0.106}
\definecolor{prompt-gray}{HTML}{a7a7a7}
\definecolor{code-syntax}{HTML}{0060b1}
\newcommand{\command}[1]{\textcolor{comment-green}{#1}}
\newcommand{\prompt}[1]{\textcolor{prompt-gray}{#1}}
\newcommand{\hlcode}[1]{\colorbox{highlight}{\makebox[0.96\linewidth][l]{#1}}}

\newcommand{\lmp}[1]{
\begin{tcolorbox}[boxsep=0pt,
                  left=3pt,
                  right=-4pt,
                  top=3pt,
                  bottom=3pt,
                  arc=0pt,
                  boxrule=0.5pt,
                  colframe=light-gray,
                  colback=white
                  ]
\small{  
\ttfamily
#1
}
\end{tcolorbox}
}

\newcommand{\speciallmp}[1]{
\begin{tcolorbox}[
 enlarge top by=0.5em,
 boxsep=0pt,
                  left=3pt,
                  right=-4pt,
                  top=3pt,
                  bottom=3pt,
                  arc=0pt,
                  boxrule=0.5pt,
                  colframe=light-gray,
                  colback=white
                  ]
\small{  
\ttfamily
#1
}
\end{tcolorbox}
}


\newpage
\setcounter{section}{0}
\setcounter{equation}{0}
\setcounter{figure}{0}
\setcounter{table}{0}
\setcounter{page}{1}
\makeatletter

\maketitle

\begin{abstract}
    \rebuttal{Recent open-vocabulary robot mapping methods enrich dense geometric maps with pre-trained visual-language features. While these maps allow for the prediction of point-wise saliency maps when queried for a certain language concept, large-scale environments and abstract queries beyond the object level still pose a considerable hurdle, ultimately limiting language-grounded robotic navigation.
In this work, we present \ours{}, a hierarchical open-vocabulary 3D scene graph mapping approach for language-grounded indoor robot navigation. Leveraging open-vocabulary vision foundation models, we first obtain state-of-the-art open-vocabulary segment-level maps in 3D and subsequently construct a 3D scene graph hierarchy consisting of floor, room, and object concepts, each enriched with open-vocabulary features. Our approach is able to represent multi-story buildings and allows robotic traversal of those using a cross-floor Voronoi graph. 
\ours{} is evaluated on three distinct datasets and surpasses previous baselines in open-vocabulary semantic accuracy on the object, room, and floor level while producing a 75\% reduction in representation size compared to dense open-vocabulary maps. In order to prove the efficacy and generalization capabilities of \ours{}, we showcase successful long-horizon language-conditioned robot navigation within real-world multi-story environments. We provide code and trial video data at: \website.}
 
\end{abstract}

\IEEEpeerreviewmaketitle


\section{Introduction}
\label{sec:introduction}
Humans acquire conceptual knowledge about the world as a whole and concrete objects in particular through multi-modal experiences. These semantic experiences are paramount to object recognition and language as well as reasoning and planning \cite{jefferies2021semantic,kumar2021semantic}. Cognitive maps store this information based on sensor fusion, fragmentation, and hierarchical structure. This is central to the human ability to navigate the physical world \cite{hirtle1985evidence,KUIPERS2000191,voicu2003hierarchical}. Recently, language proved to be an effective link between intelligent systems and humans and can enable robot autonomy in complex human-centered environments~\cite{ahn2022can, mees2023grounding, wu2023tidybot, shah2023lm, huang23vlmaps, chen2023open, shafiullah2022clipfields, conceptfusion, conceptgraphs}. 

\begin{figure}[t]
\includegraphics[width=\columnwidth]{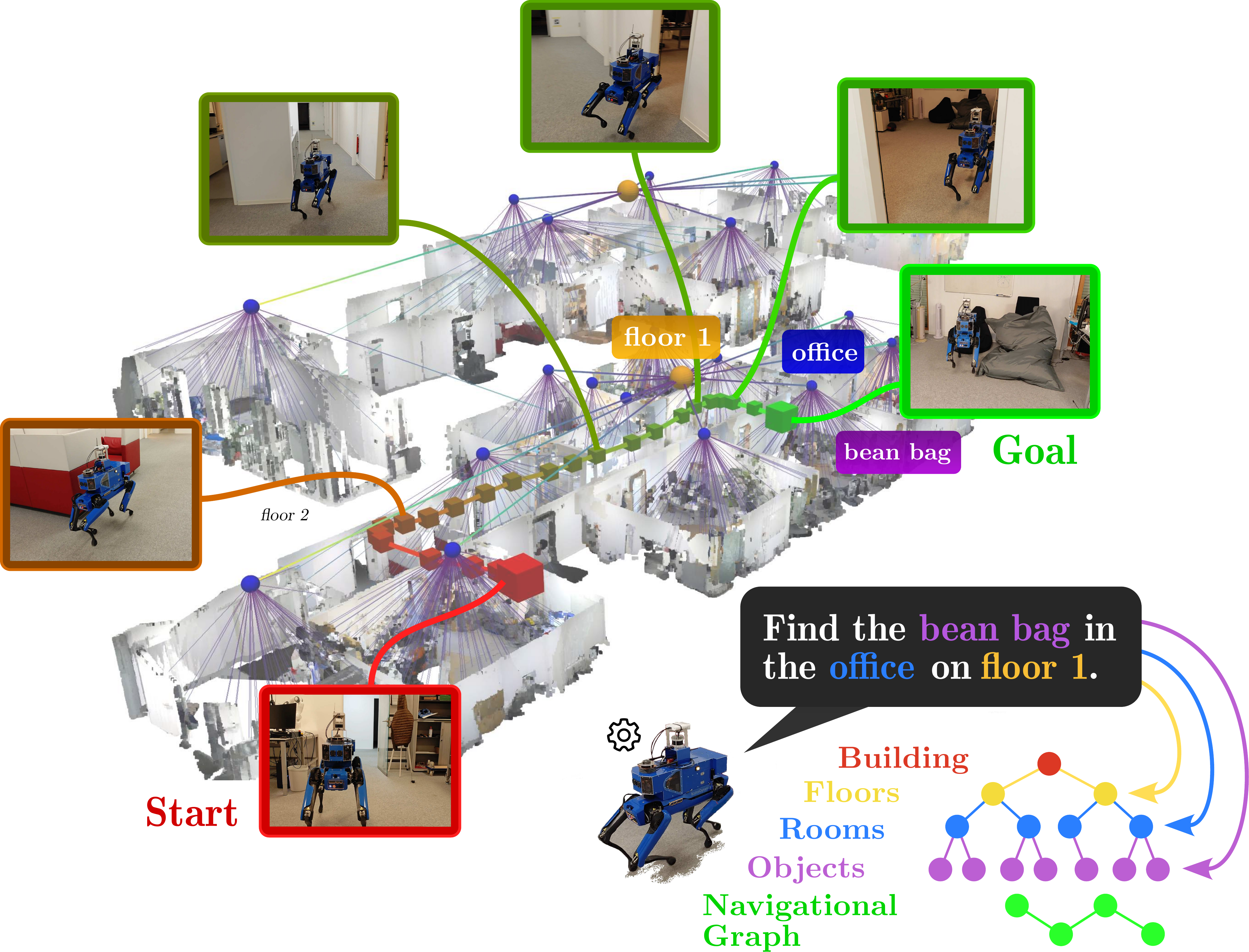}
\caption{\color{black}\ours{} enables the construction of accurate open-vocabulary 3D scene graphs for large-scale and multi-story environments and enables robots to navigate in them effectively.}
\label{fig:teaser}
\end{figure}

Classical methods for robot navigation build dense spatial maps of high accuracy using approaches to simultaneous localization and mapping (SLAM)~\cite{thrun05probabilisticrobotics}. Those give rise to fine-grained navigation and manipulation based on geometric goal specifications.
Recent advances have combined dense maps with pre-trained zero-shot vision-language models, which facilitates open-vocabulary indexing of observed environments~\cite{shah2023lm,huang23vlmaps,chen2023open,huang23avlmaps,conceptfusion, OpenScene,shafiullah2022clipfields}. While these approaches marry the area of classical robotics with modern open-vocabulary semantics, representing larger scenes while abstracting still poses a considerable hurdle. Scalable scene representations generated from real-world perception inputs should generally fulfill the following requirements: 1) Object-centricity and abstraction in terms of hierarchies, 2) efficiency regarding storage capacity as well as actionability, 3) true open-vocabulary indexing and easy querying. 

\rebuttal{
A number of works approach this using 3D scene graph structures~\cite{greve2023curb, hughes2022hydra, rosinol20203DDS} that excel at representing larger environments efficiently. At the same time, they constitute a useful interface to semantic tokens used for prompting large language models (LLM). Nonetheless, most approaches rely on closed-set semantics with the exception of ConceptGraphs~\cite{conceptgraphs} that focuses on smaller scenes.}

\color{black}
In this work, we present \textbf{H}ierarchical \textbf{O}pen-\textbf{V}ocabulary 3D \textbf{S}cene \textbf{G}raphs, short \ours{}. Our approach abstracts from dense open-vocabulary maps and allows the indexing of three distinct concepts, namely floors, rooms as well as objects. We utilize open-vocabulary vision-language models~\cite{radford2021learning,kirillov2023segany} across all concepts in order to construct 3D scene graph hierarchies that span multi-story environments while maintaining a small memory footprint. Given its concept-centric nature, the \ours{} representation is promptable using LLMs. Different from previous work, our approach relates to different conceptual levels by first decomposing abstract queries such as \textit{``towel in the bathroom on the upper floor''} and scoring the obtained tokens against the different levels of the hierarchy. We complement this with a navigational Voronoi graph that covers multiple floors including stairs, which allows actionable grounding of decomposed queries in the environment. This enables object retrieval and long-horizon robotic navigation in large-scale indoor environments from abstract queries as shown in Fig.~\ref{fig:teaser}.\color{black}

In summary, we make the following contributions:
\begin{enumerate}
    \item We introduce a novel fusion scheme using feature clustering of zero-shot embeddings that yields state-of-the-art results in open-set 3D semantic segmentation.
    \item We present an algorithm that enables the construction of truly actionable open-vocabulary 3D scene graphs of multi-floor buildings.
    \item \rebuttal{We evaluate the semantic segmentation performance of our method on the Replica~\cite{straub2019replica} and ScanNet~\cite{dai2017scannet} dataset and analyze key properties of our scene graphs on the Habitat-Matterport 3D Semantics dataset~\cite{habitat23semantics}. Furthermore, we present a detailed ablation study to justify our design choices}.
    \item We conduct real-world multi-floor object navigation experiments based on long natural language queries.
    \item We introduce a novel evaluation metric for measuring open-vocabulary semantics termed AUC$_{\text{top-k}}$.
    \item We make our code and evaluation protocol publicly available at \website{} to foster future research and introduce comparability in open-set mapping. 
\end{enumerate}

\section{Related Work}
\label{sec:related_work}

\subsection{Semantic 3D Mapping}

Enriching a geometric map with semantic information is a stepping stone to a flexible and versatile navigation system~\cite{chang2023goat,gadre2023cows,huang23avlmaps,huang23vlmaps,KUIPERS2000191,shah2023lm}. In the past, researchers created semantics-enhanced or instance-level maps by learning sensor observation features~\cite{mozos2007using}, matching pre-built object shapes to the geometric map~\cite{salas2013slam++}, back-projecting 2D semantic predictions into the 3D space~\cite{grinvald2019volumetric,mccormac2018fusion++,xu2019mid}, or instantiating 2D detections with basic 3D elements such as cubes or quadrics~\cite{nicholson2018quadricslam,yang2019cubeslam}. These methods have shown their capabilities of reconstructing scenes with both accurate geometric structure and precise semantic meaning. However, most of these methods only work with a fixed category set constrained by either the trained semantic prediction models or the pre-defined set of relevant object primitives. 

On account of recent advancements in large vision-language models such as CLIP~\cite{radford2021learning} and their fine-tuned counterparts, a number of works proposed map representations that integrate visual-language features into geometric maps, enabling open-vocabulary indexing of objects~\cite{chen2023open,huang23avlmaps,huang23vlmaps,conceptfusion,kerr2023lerf,OpenScene,shafiullah2022clipfields}, audio data~\cite{huang23avlmaps,conceptfusion} and images~\cite{chang2023goat,huang23avlmaps} in an unstructured environment. While lifting the constraints of fixed semantic categories, these approaches often necessitate the storage of a visual-language embedding for each geometric element such as points, voxels, or 2D cells in the map, resulting in a significant increase in storage overhead. 

\subsection{3D Scene Graphs}

3D scene graphs have emerged as an effective, object-centric representation of large-scale indoor \cite{armeni20193d, hughes2022hydra, rosinol20203DDS} and outdoor scenes~\cite{greve2023curb}. By representing objects or spatial concepts as nodes and their relations as edges, 3D scene graphs allow to efficiently represent larger scenes~\cite{armeni20193d,greve2023curb,hughes2022hydra}. Both edges and nodes can hold geometric and semantic attributes, which are often inferred from certain off-the-shelf networks~\cite{wu_scenegraphfusion}.
Decomposing scenes into objects and their relations enables higher-level reasoning for robotic navigation and manipulation. This is particularly useful in the realm of reasoning, planning, and navigation given the object-centric nature of these tasks~\cite{hughes2022hydra, rana2023sayplan}. Often combined with odometry estimates from simultaneous localization and mapping (SLAM)~\cite{sgraphs_2022, greve2023curb, g2o}, 3D scene graphs also allow a tight coupling between semantics and highly accurate mapping approaches utilizing e.g. meshes to represent environments~\cite{hughes2022hydra}.

Early works have shown how to encode hierarchies via abstraction in both the spatial and the semantic domain using offline approaches~\cite{armeni20193d, rosinol20203DDS}. Successive works investigated learning-based scene graph construction~\cite{wald_3dssg, wu_scenegraphfusion} as well as dynamic indoor scenes~\cite{hughes2022hydra}. Several approaches such as SceneGraphFusion~\cite{wu_scenegraphfusion} and S-Graphs~\cite{sgraphs_2022} also investigate the real-time capabilities of their proposed approaches. 
Most recently, ConceptGraphs~\cite{conceptgraphs} was the first to show how to combine 3D scene graphs with open-vocabulary vision-language features. In addition, the authors show how to query the graph using LLMs and demonstrate \rebuttal{various downstream applications}. 

\begin{figure*}[t]
\includegraphics[width=\textwidth]{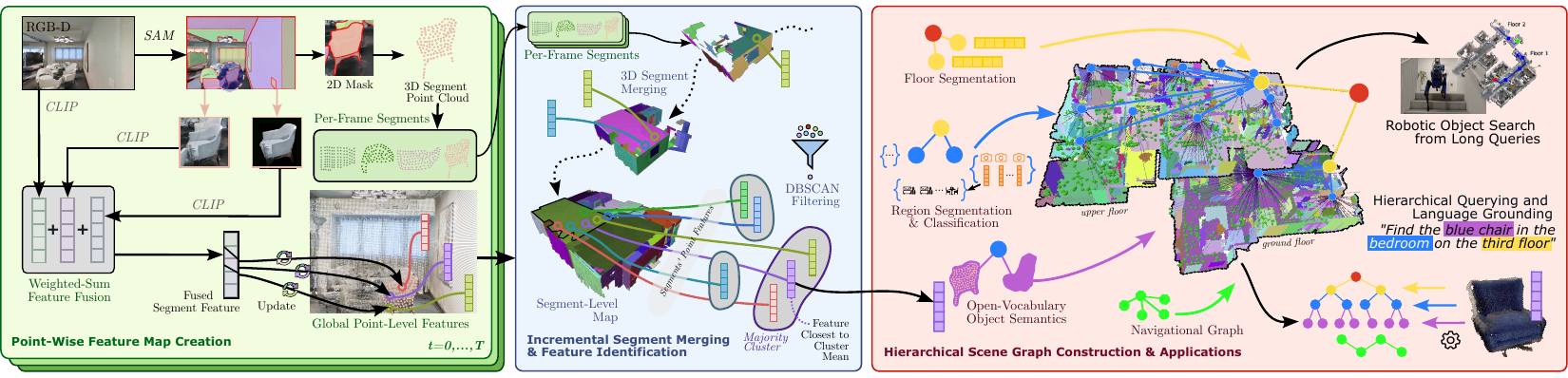}
\caption{\color{black} HOV-SG builds hierarchical open-vocabulary 3D scene graphs of indoor household scenes. We first use SAM to extract object masks per frame while obtaining vision-language features via CLIP. In the next step, we aggregate these features on a point level in the map. Secondly, we segment the full point cloud based on merged 3D masks. To generate more meaningful semantic object features, we employ a DBSCAN-based filtering approach to obtain a majority vote feature for each object. To construct an actionable 3D scene graph, we segment the obtained panoptic map into multiple floors, segment and classify distinct regions using several view embeddings, and identify object names via querying. As a result, HOV-SG allows hierarchical querying and navigation using mobile robots even in complex multi-floor environments.}
\label{fig:overview}
\end{figure*}

\subsection{Scene Graphs for Planning}
\label{sec:scene-graph-planning}
\color{black} 
Several recent works have investigated the use of scene graphs for robotic planning. The earliest approaches rely on pre-explored environments and perform iterative scene graph decomposition to retrieve grounded plans~\cite{rana2023sayplan, chalvatzaki2023learning}. 
RobLM~\cite{chalvatzaki2023learning} decomposes the planning stage by relying on a fine-tuned GPT-2 instance that proposes high-level sub-problems from scene graphs, which are in turn solved through PDDL task planners. SayPlan~\cite{rana2023sayplan} directly utilizes GPT-4~\cite{openai2023gpt} for iterative search on a scene graph to generate grounded plans, which requires feasibility constraints on the manipulated entities and actions. Another line of work investigates robotic navigation from scene graphs. SayNav~\cite{rajvanshi2023saynav} obtains LLM-generated plans from scene graphs and executes short-distance point-goal navigation sub-tasks. Contrary to that, VoroNav~\cite{wu2024voronav} constructs a Voronoi graph that is attributed to camera observations in order to solve object navigation. Orthogonal to that, MoMa-LLM~\cite{honerkamp2024language} tackles mobile manipulation objectives using scene graphs fed to GPT-4 in a task-specific manner. Similarly, GRID~\cite{ni2023grid} uses a graph neural network to predict actions from scene graphs and LLM encodings.
\color{black}

\vspace{1em}
Conceptually, our work is most similar to ConceptGraphs and Hydra. While ConceptGraphs is only evaluated on small scenes and mostly validated by human evaluators in terms of semantic accuracy of nodes etc., our work proposes not only a novel metric for measuring the semantic accuracy of object features but also introduces open-vocabulary hierarchies. Different from ConceptGraphs and similar to Hydra~\cite{hughes2022hydra}, which does not operate on open-vocabulary features, our approach demonstrates how to efficiently represent actionable, hierarchical 3D scene graphs that are attributed with open-vocabulary features.








\section{Technical Approach}
\label{sec:approach}

\rebuttal{This work aims to develop a concise and efficient visual-language graph representation for large-scale multi-floor indoor environments given RGB-D observations and odometry. The graph should facilitate the indexing of multi-level semantic concepts through natural language queries} such as ``the first floor'' (floor level), ``the office on the first floor'' (room level), and ``the plant in the office on the second floor'' (object level). Additionally, the graph should be actionable and enable a robot to localize and navigate semantically and spatially in the environment without additional geometric maps. \rebuttal{We address this by introducing} \textbf{H}ierarchical \textbf{O}pen-\textbf{V}ocabulary \textbf{S}cene \textbf{G}raphs, \rebuttal{in short \ours{}}. 
The overall pipeline consists of two stages. We first create a 3D segment-level open-vocabulary map and then build a hierarchical open-vocabulary scene graph based on the map. In the following sections, we describe (i) the construction of the 3D segment-level open-vocabulary map (Sec.~\ref{subsec:segment_map_creation}), (ii) the creation of the hierarchical open-vocabulary scene graph (Sec.~\ref{subsec:scene_graph_creation}), and (iii) how to use the graph for language-conditioned navigation across a large-scale environment (Sec.~\ref{subsec:scene_graph_navigation}). Fig.~\ref{fig:overview} presents an overview of our method.

\subsection{3D Segment-Level Open-Vocabulary Mapping}
\label{subsec:segment_map_creation}


The main idea of building a segment-level open-vocabulary map is to create a list of 3D point clouds, namely segments, from an RGB-D video with odometry and assign an open-vocabulary feature generated by a pre-trained visual-and-language model (VLM) to each segment. Unlike previous works that equip each 3D point with an independent visual-language feature~\cite{chen2023open,huang23vlmaps,conceptfusion,shafiullah2022clipfields}, we leverage the fact that neighboring points in the 3D world often share the same semantic information. \rebuttal{This implies the potential of reducing the required semantic features to represent the scene while maintaining expressiveness.}

\begin{figure*}[ht]
\includegraphics[width=\textwidth]{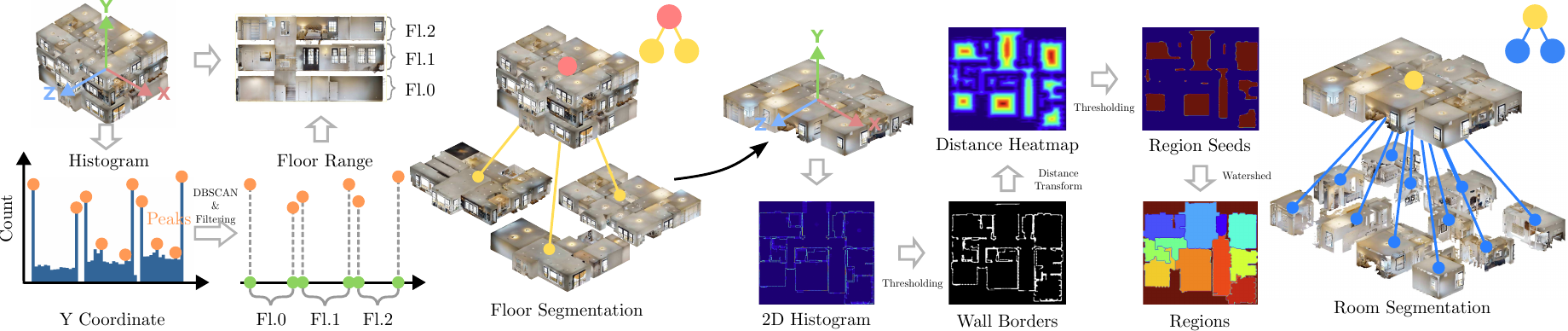}
\caption{Floor and Room Segmentation. Given the point cloud of the whole environment, floor and room nodes are subsequently derived based on geometric heuristics. Floor boundaries are computed by finding peaks of the pixel density along the height direction followed by filtering while room segment masks are extracted using the Watershed algorithm.}
\label{fig:floor_room_seg}
\end{figure*}

\noindent\textbf{Frame-Wise 3D Segment Merging}: 
\label{subsubsec:frame_wise_segment_creation}
\rebuttal{Given a sequence of RGB-D observations, we utilize Segment Anything~\cite{kirillov2023segany} to obtain a list of class-agnostic 2D binary masks at each timestep. The pixels in each mask are then backprojected to 3D using the depth information, resulting in a list of point clouds, or 3D segments. Based on accurate odometry estimates, we transform all 3D segments into the global coordinate frame. These frame-wise segments are either initialized as new global segments or merged with existing ones based on an overlap metric:}
\begin{equation}
    R(m, n) = max(\textit{overlap}(S_m, S_n), \textit{overlap}(S_n, S_m)),
\end{equation}
\rebuttal{where $S_m$ and $S_n$ indicate segment (or point cloud) $m$ and $n$, $\textit{overlap}(S_a, S_b)$ is computed by taking the number of points in $S_a$ showing a neighbor in $S_b$ within a certain distance divided by the total number of points in $S_a$. Different from Gu~\emph{et al.}~\cite{conceptgraphs}, who incrementally merge new segments with one global segment that has the largest overlapping ratio, we construct a fully connected graph where each segment serves as a node and their edge weights are the corresponding overlapping ratios. Based on these weights, highly-connected subgraphs are subsequently merged. In this way, one segment can be merged with multiple segments, which is useful in situations in which an incoming segment is, e.g., filling a gap between two already registered global segments.}

\label{subsubsec:segment_features_creation}
\noindent\textbf{Segment-Level Open-Vocabulary Features Computation}:
\rebuttal{For each obtained 2D SAM mask per frame, we obtain an image crop based on its bounding box as well as an image of the isolated mask without background. We encode the full RGB frame and the two mask-wise images with CLIP~\cite{radford2021learning} and fuse them in a weighted-sum manner (Fig.~\ref{fig:overview}, left). Previous work~\cite{liang2023open} proposed to use the CLIP feature of the masked image without background, while others~\cite{conceptfusion} approach this by combining the CLIP features of the whole image and the target mask's crop including background. In our work, we empirically show that encoding the full RGB frame and the two mask-wise images with CLIP and fusing them in a weighted-sum manner achieves improved results (Sec.~\ref{sec:exp_ablation}). The fusion scheme can be formulated as:}
\begin{equation}
    f_i = w_g f_g + w_l f_l + w_m f_m,
\end{equation}
\rebuttal{where $f_i$ indicates the fused features for the $i$-th 2D mask in the frame, $f_g$, $f_l$, and $f_m$ indicate the CLIP features extracted from the entire RGB frame, the image crop of the 2D mask, and the image crop of the 2D mask excluding the background, respectively. Furthermore, $w_g$, $w_l$, and $w_m$ represent their respective weights, which sum up to 1.}

\rebuttal{Assuming a single CLIP feature for each mask, we transform the 2D mask into global 3D coordinates and associate the obtained fused CLIP feature with the nearest 3D points in a pre-computed reference point cloud. Based on this association, we register the obtained segment features on a global point-wise feature map. The final point-wise features are determined by averaging each reference point's associated features. Based on the 3D segments obtained in the independent merging step, we can finally infer open-vocabulary vision-language features for all 3D segments as outlined in Fig.\ref{fig:overview}. 
In the subsequent step, we match point-wise features with the obtained 3D segments. For each point within a segment, we identify the nearest points in the reference point cloud and collect their CLIP features. We leverage DBSCAN to cluster all the point-wise features of the segment and assign the feature that is closest to the majority cluster’s mean to the segment (Fig.~\ref{fig:overview}, middle). This circumvents mode collapse while removing noise and thus produces more semantically meaningful segment features.}

\subsection{3D Scene Graph Construction}
\label{subsec:scene_graph_creation}
\rebuttal{In this section, we describe how to build a hierarchical open-vocabulary scene graph given a global reference point cloud of the scene, a list of global 3D segments, and their associated CLIP features as described in Sec.~\ref{subsec:segment_map_creation}.}

\rebuttal{We formalize our graph as $\mathcal{G} = (\mathcal{N}, \mathcal{E})$ where $\mathcal{N}$ denotes the nodes and $\mathcal{E}$ denotes the edges. The nodes can be expressed as $\mathcal{N} = \mathcal{N}_S \cup \mathcal{N}_F \cup \mathcal{N}_R \cup \mathcal{N}_O$, consisting of a root node $\mathcal{N}_S$, floor nodes $\mathcal{N}_F$, room nodes $\mathcal{N}_R$, and object nodes $\mathcal{N}_O$. Each node in the graph except the root node $\mathcal{N}_{S}$ contains the point cloud of the concept it refers to and the open-vocabulary features associated with it. The edges can be written as $\mathcal{E} = \mathcal{E}_{SF} \cup \mathcal{E}_{FR} \cup \mathcal{E}_{RO}$. Here, $\mathcal{E}_{SF}$ represents the edges between the root node and the floor nodes, $\mathcal{E}_{FR}$ represents the edges between the floor nodes and the room nodes, and lastly, $\mathcal{E}_{RO}$ denotes the edges between the room and object nodes.}

\noindent\textbf{Floor Segmentation}:
\label{subsubsec:floor_seg}
\rebuttal{In order to separate floors, we identify peaks of a height histogram over all points contained in the point cloud. 
Given the point cloud of the whole environment, we construct the histogram over all points along the height axis using a bin size \SI{0.01}{\meter}. Next, we identify peaks in this histogram (within a local range of \SI{0.2}{\meter}) and select only peaks that exceed a minimum of 90\% of the highest intensity peak. We apply DBSCAN and select the two highest-ranking peaks in each cluster. After that, every two consecutive values in the sorted height vector represent a single floor (floor and ceiling) in the building. The floor segmentation process is shown in Fig.~\ref{fig:floor_room_seg}.
Using the obtained height levels, we can extract floor point clouds for each floor $\mathcal{P}_{l}$ where $l$ is the floor number. In addition, we equip each floor node with a CLIP text embedding using the template ``floor \{\#\}''. A graph edge between the root node and each floor node $(N_S, N_{l}) \in \mathcal{E}_{SF}$ is established. }

\noindent{\textbf{Room Segmentation:}
\label{subsubsec:room_seg}
\rebuttal{Based on each obtained floor point cloud, we construct a 2D bird's-eye-view (BEV) histogram, from which a binary wall skeleton mask is extracted by thresholding the histogram. After dilating the wall mask and computing an Euclidean distance field (EDF), a number of isolated regions is derived by thresholding the EDF. Taking these regions as seeds, we apply the Watershed algorithm to obtain 2D region masks. The room segmentation process is further shown in Fig.~\ref{fig:floor_room_seg}.
Given the 2D region masks, we extract the 3D points that fall into the floor's height interval as well as the BEV room segment to form room point clouds that are used to associate objects to rooms later.}

\rebuttal{To enrich room nodes with open-vocabulary features, we associate RGB-D observations whose camera poses reside within a room segment to those rooms (see Fig.~\ref{fig:room_embedding}). The CLIP embeddings of these images are distilled by extracting $k$ representative view embeddings using the k-means algorithm. During inference, given a list of room categories encoded via CLIP, we construct a cosine similarity matrix between the $k$ representative features and all room category features. 
Next, we take the $\operatorname{argmax}$ along the category axis and obtain the most probable room type for each representative separately, resulting in $k$ votes per room. Given these votes, we obtain the predicted room category by either taking the maximum-score vote or the majority vote across all $k$ representatives per room.}

\rebuttal{These $k$ representative embeddings and the room point cloud attribute the room node $N_{f,r}$ of room $r$ on floor $f$. An edge between the floor node and each room node $(N_{f}, N_{f,r}) \in \mathcal{E}_{FR}$ is established. The construction and querying of view embeddings are illustrated in Fig.~\ref{fig:room_embedding}.}

\begin{figure}
\includegraphics[width=0.49\textwidth]{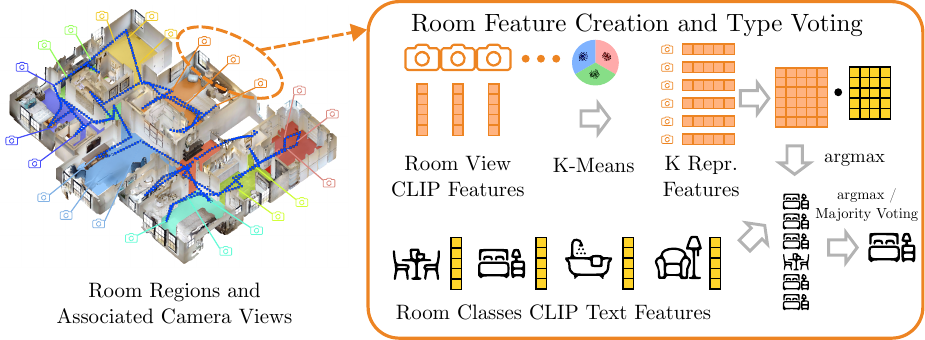}
\caption{Room embedding computation and room type voting. We enrich each room node with open-vocabulary embeddings by associating the observations with it. Given the segmented room region and the contained camera poses we extract 10 distinct CLIP features that represent the semantic distribution of a room.}
\label{fig:room_embedding}
\end{figure}

\noindent\textbf{Object Identification}:
\label{subsubsec:object_iden}
\rebuttal{Given the room point cloud, we associate object-level 3D segments that show a point cloud overlap with a potential candidate room in the bird's-eye-view. Whenever a segment shows zero overlap with any room, we associate it with the room showing the smallest Euclidean distance. To reduce the number of nodes, we merge 3D segments of significant pair-wise partial overlap (Sec.~\ref{subsubsec:frame_wise_segment_creation}) that produce equal object labels when queried against a chosen label set. Each merged point cloud constitutes an object node $N_{f,r,o}$ that is connected to its corresponding room node $N_{f,r} \in \mathcal{N}_{R}$ by an edge $(N_{f,r}, N_{f,r,o}) \in \mathcal{E}_{RO}$. Each object node holds its corresponding 3D segment feature as described in Sec.~\ref{subsubsec:segment_features_creation}, its 3D segment point cloud as well as a maximum-score object label for intermediate naming.}


\begin{figure}[t]
    \centering
    \begin{subfigure}{\linewidth}
        \centering
        \includegraphics[width=\linewidth]{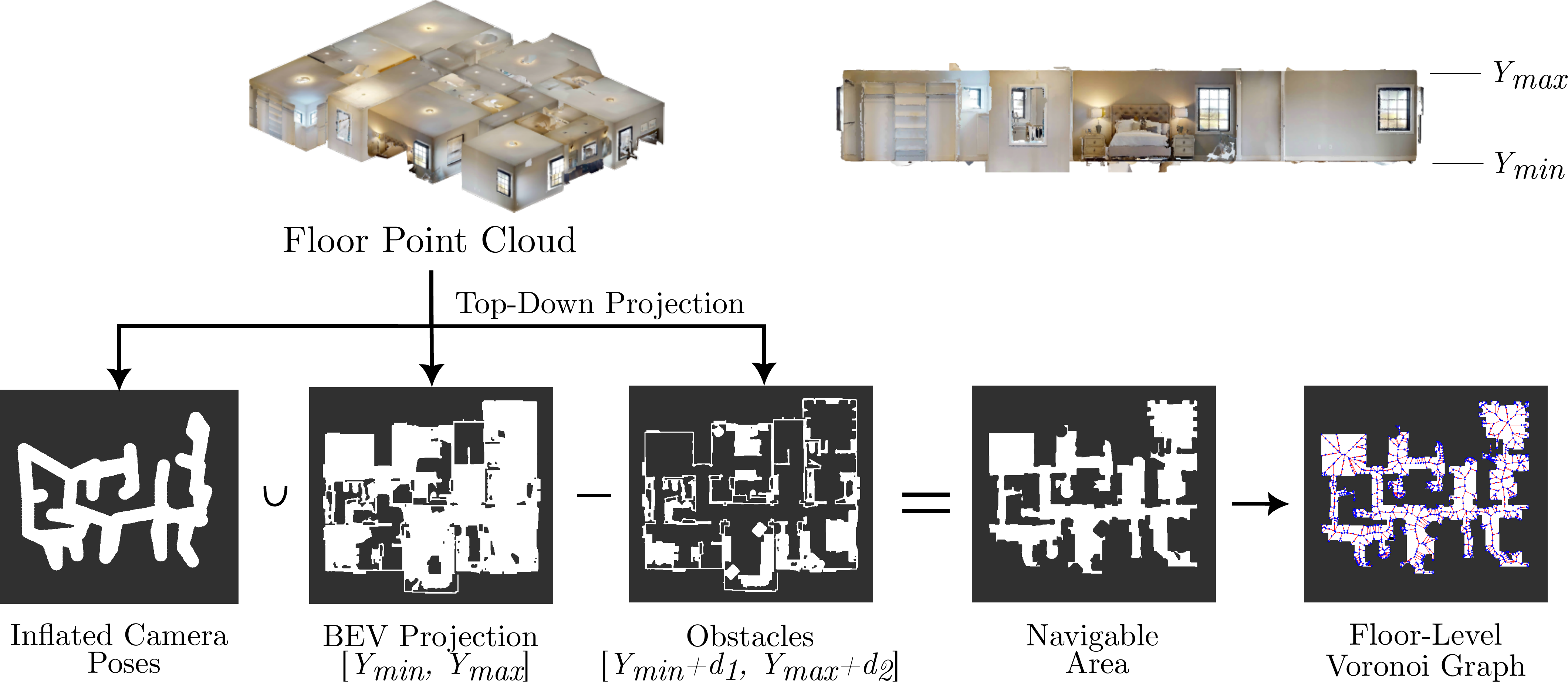}
        \caption{Single-floor Navigational Graph}
        \label{fig:floor_level_voronoi}
    \end{subfigure}
    \begin{subfigure}{\linewidth}
        \centering
        \includegraphics[width=\linewidth]{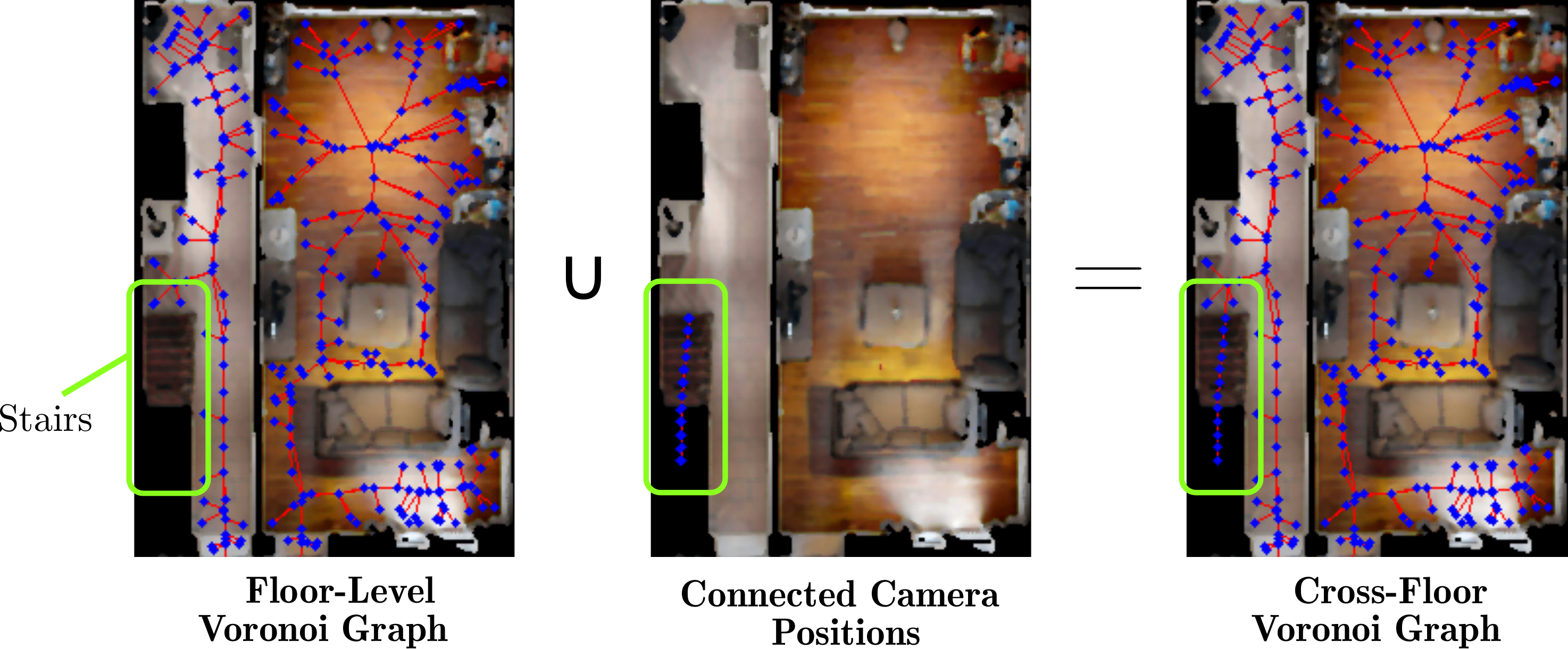}
        \caption{Cross-floor Navigational Graph}
        \label{fig:cross_floor_voronoi}
    \end{subfigure}
    \caption{\rebuttal{Actionable navigational graph: The creation of the actionable navigational graph involves constructing single-floor and cross-floor navigational graphs: (a) By deducting the set of obstacles from the union of camera poses and the per-floor obtained BEV projection of the floor point cloud, we obtain the navigable area. Within this area we construct a Voronoi diagram as shown right. (b) In order to equip our navigational graph with cross-floor navigation capabilities, we extract the camera positions within regions classified as \textit{stairs}. This subgraph is connected with the corresponding floor-level Voronoi graphs.}}
    \label{fig:whole}
\end{figure}

\noindent\textbf{Actionable Graph Creation}:
\rebuttal{In addition to the open-vocabulary hierarchy, the scene graph also contains a navigational Voronoi graph that serves robotic traversability of the mapped surroundings~\cite{Thr96Int} spanning multiple floors. This enables high-level planning and low-level execution based on the Voronoi graph. The creation of actionable graphs involves constructing per-floor and cross-floor navigation graphs. For the floor-level graph, the approach entails computing the free space map of the floor and creating a Voronoi graph~\cite{Thr96Int} based on it. To construct per-floor graphs, we first obtain all camera poses and project them as 2D points onto a BEV map of each floor, assuming areas within a certain radius of two nodes are pair-wise navigable. Subsequently, the entire floor's region is obtained by projecting all floor-wise points into the BEV plane. An obstacle map is generated based on points within a predefined height range $[y_{min} + \delta_1, y_{min} + \delta_2]$, where $y_{min}$ is the minimal height of the floor points while $\delta_1 = 0.2$, $\delta_2 = 1.5$ are empirically tuned. By taking the union of the pose region map and the floor region map and subtracting the obstacle region map, the free space map of each floor is derived. The Voronoi graph of this free map yields the floor graph. (see Fig.~\ref{fig:floor_level_voronoi}). To build cross-floor navigational graphs, camera poses on stairs are connected to form stair-wise graphs. Subsequently, the closest nodes between the stairs graph and the floor-wise graph are selected respectively and connected, thereby completing the construction of cross-floor navigational graphs as shown in Fig.~\ref{fig:cross_floor_voronoi}.}

\subsection{Navigation with Scene Graph}
\label{subsec:scene_graph_navigation}

\rebuttal{
\ours{} extends the scope of potential navigation goals to more specific spatial concepts such as regions and floors compared to simple object goals~\cite{conceptgraphs, huang23avlmaps, huang23vlmaps,conceptfusion}. Language-guided navigation with \ours{} involves processing complex queries such as \textit{``find the toilet in the bathroom on floor 2''} using a large language model (prompts are given in the supplementary material Sec.~\ref{subsec:prompt-supp}). We break down such lengthy instructions into three separate queries: one for the floor level, the room level, and the object level, respectively. Leveraging the explicit hierarchical structure of \ours{}, we sequentially query against each hierarchy level to progressively narrow down the solution corridor. This is done by taking the cosine similarity between the identified query floor, query region, and query object as well as all objects, rooms, and floors given in the graph, respectively. Once a target node is identified via scoring, we utilize the navigational graph mentioned above to plan a path from the starting pose to the target destination, which is demonstrated in Fig.~\ref{fig:long_query_nav_approach} and visualized in Fig.~\ref{fig:teaser}.
}


\section{Experimental Evaluation}
\label{sec:experiments}
The goals of our experiments are five-fold: (i) we quantitatively compare HOV-SG with recent open-vocabulary map representations in 3D semantic segmentation on ScanNet and Replica (Sec.~\ref{sec:exp_replica_scannet}), (ii) we investigate the semantic and geometric accuracy of HOV-SG at the floor, room, and object level on the Habitat Matterport 3D Semantic Dataset (Sec.~\ref{sec:exp_graph_eval_hm3dsem}), (iii) we study how HOV-SG enables large-scale language-grounded navigation in the real-world (Sec.~\ref{sec:exp_lang_nav_real_world}), (iv) we demonstrate the compact memory footprint of HOV-SG compared to previous open-vocabulary representations (Sec.~\ref{sec:exp_size}), and lastly, (v) we justify our design choices through an ablation study (Sec.~\ref{sec:exp_ablation}).

\begin{figure*}[t]
\centering
\includegraphics[width=0.8\textwidth]{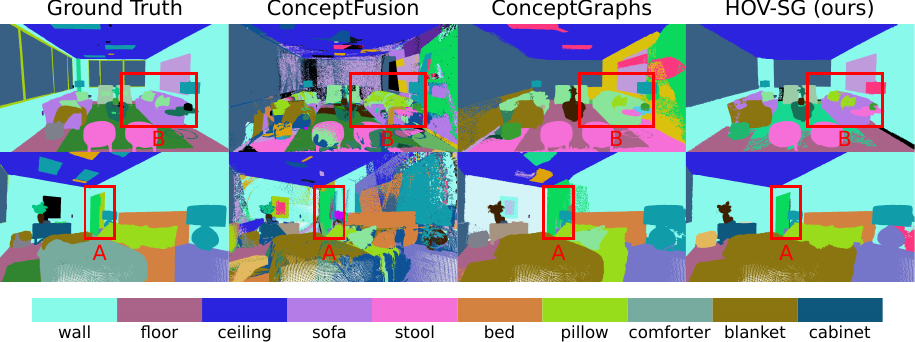}
\caption{\rebuttal{Qualitative results for 3D semantic segmentation on Replica dataset. For conciseness, the legend only shows ten out of 101 categories in the Replica dataset. Observing the position ``A'' indicated in the images, only \ours{} predicts the door with correct boundaries. At position ``B'' in the images, only \ours{} predicts the sofa correctly.}}
\label{fig:replica_qualitative}
\end{figure*}

\begin{table}[t]
\centering
\scriptsize
\setlength\tabcolsep{2.5pt}
\caption{\textsc{Open-Vocabulary 3D Semantic Segmentation}}
\begin{threeparttable}
    \begin{tabular}{lr|ccc|ccc}
     \toprule
     \multirow{2}{*}{Method} & CLIP &
     \multicolumn{3}{c|}{Replica} & \multicolumn{3}{c}{ScanNet} \\
     
    & Backbone & mIOU & F-mIOU & mAcc & mIOU & F-mIOU & mAcc \\
    
     \midrule
     \rowcolor{black!10} 
     \color{black}MinkowskiNet~\cite{choy20194d}  &  & - & - & - & \color{black} 0.42 & \color{black} 0.47 & \color{black} 0.56 \\
     \greyrule
    \multirow{2}{*}{ConceptFusion~\cite{conceptfusion}}  & OVSeg & 0.10 & 0.21 & 0.16 & 0.08 & 0.11 & 0.15 \\
     & Vit-H-14 & 0.10 & 0.18 & 0.17 & 0.11 & 0.12 & 0.21 \\
     \greyrule
    \multirow{2}{*}{ConceptGraph~\cite{conceptgraphs}}  & OVSeg & 0.13 & 0.27 & 0.21 & 0.15 & 0.18 & 0.23 \\
   
     & Vit-H-14 & 0.18 & 0.23 & 0.30 & 0.16 & 0.20 & 0.28 \\
    
    \midrule    
    \multirow{2}{*}{HOV-SG (ours)} & OVseg & 0.144 & 0.255 & 0.212 & 0.214 & 0.258 & 0.420 \\
     & Vit-H-14 & \textbf{0.231} & \textbf{0.386} & \textbf{0.304} & \textbf{0.222 }& \textbf{0.303} & \textbf{0.431} \\

    \bottomrule
    \end{tabular}
    Higher values are better. \rebuttal{The used evaluation metrics are defined in Sec.~\ref{sec:evaluation-metrics-3d-seg-suppl} in the supplementary material.} The ConceptFusion pipeline evaluated against made use of instance masks predicted by SAM~\cite{kirillov2023segany}. \color{black} The MinkowskiNet~\cite{choy20194d} is a privileged method that was trained on the full set of ScanNet~\cite{dai2017scannet} scenes to demonstrate the gap between zero-shot and fully-supervised methods.
\end{threeparttable}
\label{tab:seg-comparison}
\end{table}

\subsection{3D Semantic Segmentation on ScanNet and Replica}
\label{sec:exp_replica_scannet}
To test the semantic expressiveness of our \ours{} method, we evaluate the open-vocabulary 3D semantic segmentation performance on ScanNet~\cite{dai2017scannet} and Replica~\cite{straub2019replica}. We compare our method with two alternative vision-language representations (ConceptFusion~\cite{conceptfusion} and ConceptGraphs~\cite{conceptgraphs}) while using different CLIP backbones. We consider ViT-H-14 and a fine-tuned backbone ViT-L-14 released with the work OVSeg~\cite{liang2023open}. \rebuttal{The used evaluation metrics are defined in Sec.~\ref{sec:evaluation-metrics-3d-seg-suppl}. To demonstrate the existing gap between zero-shot and fully supervised methods we also evaluated a MinkowskiNet~\cite{choy20194d} instance trained on ScanNet~\cite{dai2017scannet}.}

\noindent\textbf{Prediction Generation}: \rebuttal{We generate the CLIP text embedding for each category contained in the dataset by using a template of the form ``There is the \{category\} in the scene.'' as well as the category name ``\{category\}'' itself. Next, we average the two to obtain the embedding of each specific category. We obtain predicted labels for each object node by computing the cosine similarity between all object nodes' embeddings and all category embeddings and lastly apply the $\operatorname{argmax}$ operator. In the following, we concatenate all objects' point clouds to create our predicted point cloud $\mathcal{P}_{pred}$ and transform it to the same coordinate frame as that of the point cloud with ground-truth (GT) semantic labels $\mathcal{P}_{GT}$. 
Given that the predicted point cloud may exhibit varying point densities compared to the ground truth (GT), we iterate through each GT point to locate its five nearest points in $\mathcal{P}_{pred}$, and then determine the majority label among these points as the predicted label for each GT point.}

\noindent\textbf{Evaluation Scenes}: For consistency, we evaluate the same scenes evaluated in~\cite{conceptgraphs,conceptfusion}. For ScanNet, we evaluate scenes: \texttt{scene0011\_00, scene0050\_00, scene0231\_00, scene0378\_00, scene0518\_00}. Regarding Replica, we evaluate on \texttt{office0-office4} and \texttt{room0-room2}.

\noindent\textbf{Results}: The semantic segmentation results on both Replica and ScanNet are provided in Table~\ref{tab:seg-comparison}. Regarding mIOU and F-mIOU, \ours{} outperforms the open-vocabulary baselines by a large margin. This is primarily due to the following improvements we made: First, when we merge segment features, we consider all point-wise features that each segment covers and use DBSCAN to obtain the dominant feature, which increases the robustness compared to taking the mean as done by ConceptGraphs~\cite{conceptgraphs}. Secondly, when we generate the point-wise features, we use the mask feature which is the weighted sum of the sub-image and its contextless counterpart. This mitigates the impact of salient background objects. Further qualitative results are given in Fig.~\ref{fig:replica_qualitative}.

\subsection{Scene Graph Evaluation on Habitat 3D Semantics}
\label{sec:exp_graph_eval_hm3dsem}

\rebuttal{We evaluate our scene graph on four aspects. To analyze the geometric accuracy of the scene graph, we analyze the floor and class-agnostic region segmentation performance in Sec.~\ref{sec:exp-room-floor-segmentation}. To evaluate the semantic accuracy, we evaluate predicted region semantics (Sec.~\ref{sec:exp-semantic-room-classification}) as well as open-vocabulary object-level semantics (Sec.~\ref{sec:exp-hm3dsem-objects}). To scrutinize the downstream navigation capabilities of \ours{}, we conduct hierarchical object retrieval and navigation experiments given abstract language queries in Sec.~\ref{sec:exp-hier-concept-retrieval}. We display two exemplary constructed hierarchical 3D scene graphs in Fig.~\ref{fig:hm3dsem-sg-vis}. The 3D scene graph visualization of the remaining scenes is shown in Fig~\ref{fig:hm3dsem-sg-vis-suppl}.
}

\noindent\rebuttal{
\textbf{Dataset:} In order to evaluate various aspects of the produced scene graph hierarchy, we have chosen the Habitat-Semantics dataset (HM3DSem) as it provides true open-vocabulary labels across large multi-floor scenes and also provides object-region assignments. Since our approach operates on RGB-D frames, we manually record random walks of 8 scenes of the Habitat Semantics dataset~\cite{habitat23semantics}, which span multiple rooms and floors: \texttt{00824}, \texttt{00829}, \texttt{00843}, \texttt{00861}, \texttt{00862}, \texttt{00873},  \texttt{00877}, \texttt{00890}. To construct ground-truth maps to compare against, we fuse the RGB-D and panoptic data across all frames given accurate odometry and obtain RGB and panoptic global ground truth. These maps are finally voxelized using a \SI{0.02}{cm} resolution.}

\vspace{0.2cm}

\subsubsection{Floor and Class-Agnostic Room Segmentation}
\label{sec:exp-room-floor-segmentation}
In order to evaluate both floor and class-agnostic room segmentation, we identified several heuristics on top of the provided metadata of the dataset. Since HM3DSem does not include floor height labels, we hand-labeled all upper and lower floor boundaries. In addition, we pooled the annotated objects contained in our constructed ground truth point clouds based on their associated region labels. Based on this, we obtain region-wise point clouds we utilize as ground truth. \rebuttal{As shown in Table~\ref{tab:floor-region}, our method achieves 100\% accuracy in retrieving the number of floors, both in single-floor as well as multi-floor scenes. In addition, we evaluate the region segmentation performance of \ours{} and compare with Hydra~\cite{hughes2022hydra} across eight scenes on HM3DSem. We observe slightly lower precision but a significantly greater recall of 83.59\% compared to 77.55\%. In addition to the overall results given in Table~\ref{tab:floor-region}, we provide scene-wise results in Table~\ref{tab:floor-region-suppl} in the supplementary material.} 
In general, we obtain higher precision and recall on smaller scenes comprising fewer regions. Similar to Hydra~\cite{hughes2022hydra}, our approach utilizes a na\"ive morphological heuristic to segregate regions, which does not work well on more complex, semantically ambiguous room layouts such as combined kitchen and living rooms. Nonetheless, our approach does not suffer from this drawback too drastically as we equip each segmented region with 10 representative embeddings. This allows adaptive prompting without directly setting a fixed room category.

\begin{table}[t]
\color{black}
\centering
\footnotesize
\caption{\textsc{Floor and Region Segmentation on HM3DSem}}
\setlength\tabcolsep{10.7pt}
\begin{threeparttable}
\begin{tabular}{l|c|cc}
 \toprule
\multirow{2}{*}{Method} & Floors & \multicolumn{2}{c}{Regions} \\
&  $\text{Acc}_{F}$\,[\%] & $\text{Precision}$\,[\%] &  $\text{Recall}$\,[\%] \\
 \midrule
Hydra \cite{hughes2022hydra} & - & \textbf{86.18} & 77.55 \\
HOV-SG (ours) & 100 & 84.10 & \textbf{83.59} \\
\bottomrule
\end{tabular}
\footnotesize
Evaluation of the floor and room segmentation: We present the accuracy of correctly predicted floors using a threshold of $\SI{0.5}{\meter}$. The region segmentation precision (P) and recall (R) are calculated based on the metric in Hydra~\cite{hughes2022hydra}.
\end{threeparttable}
\label{tab:floor-region}
\color{black}
\end{table}
\vspace{0.2cm}

\begin{figure*}
\footnotesize
\setlength{\tabcolsep}{0.1cm}
\begin{tabular}{ccc}
{\centering\includegraphics[width=0.55\linewidth]{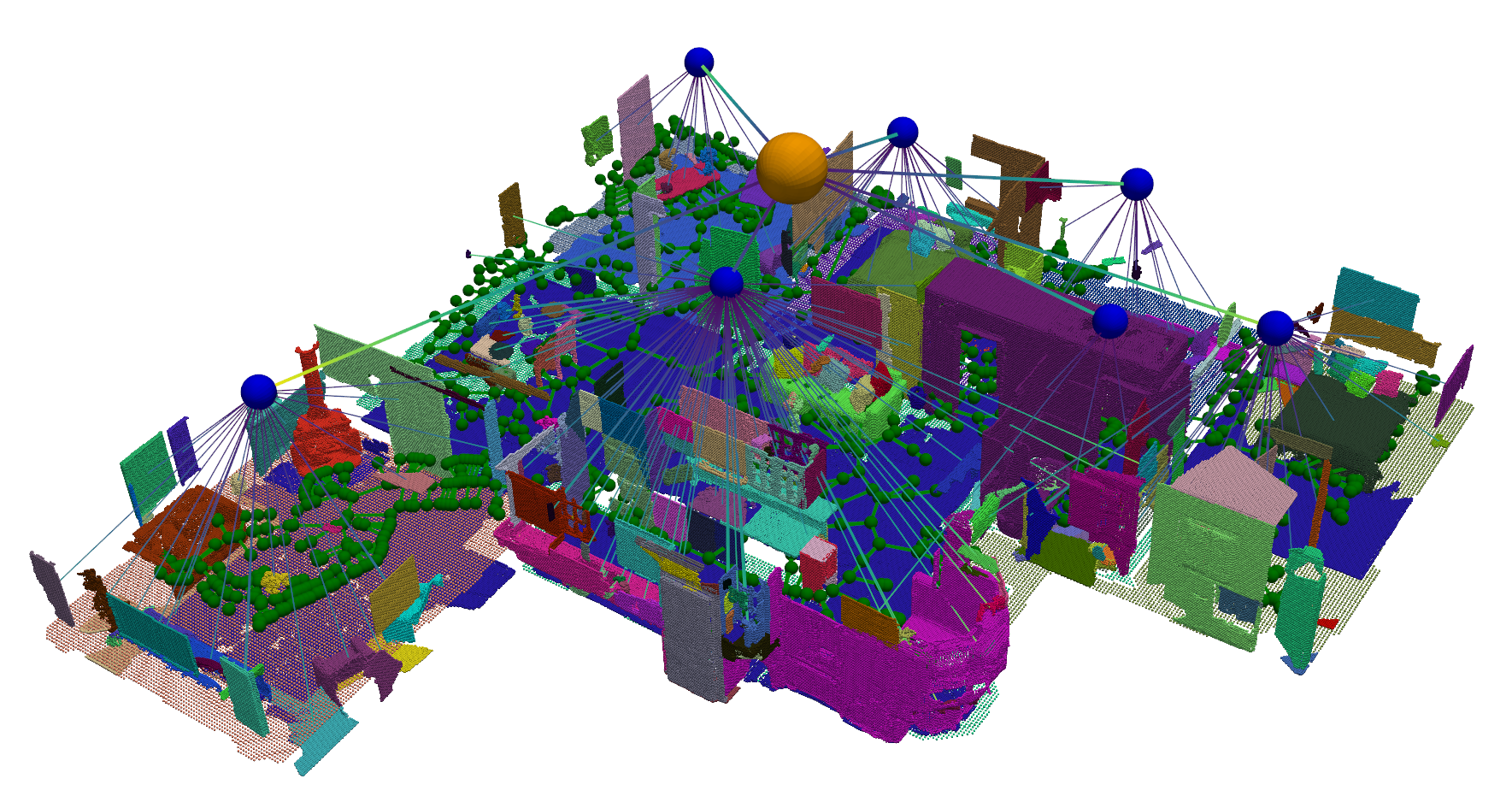}} & &
{\centering\includegraphics[width=0.4\linewidth]{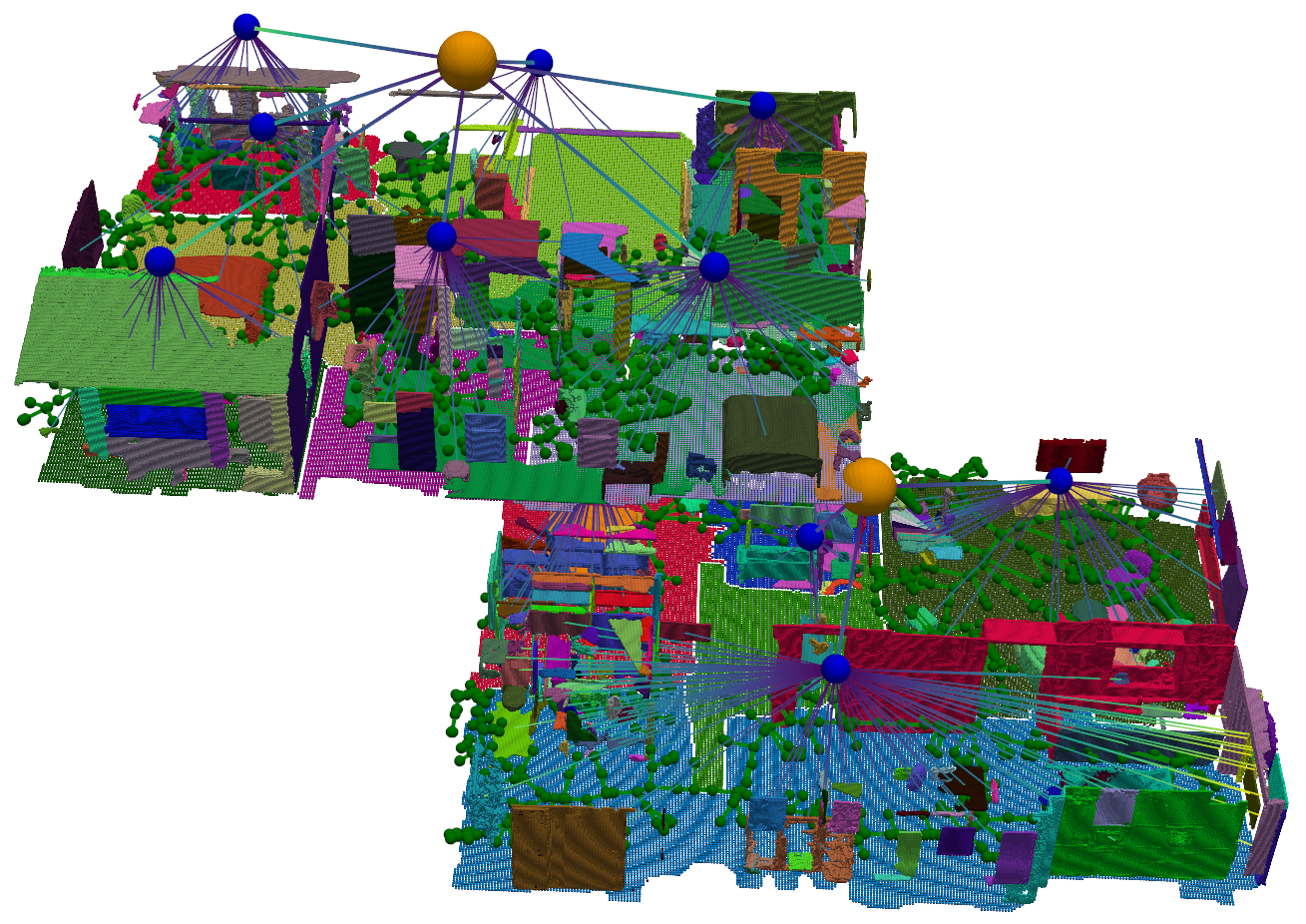}} \\
(a) \textit{Scene 00824 (single floor)} & &(b) \textit{Scene 00861 (two floors)} \\
\end{tabular}
\caption{\rebuttal{Qualitative visualization of the hierarchical 3D scene graphs produced by \ours{} of two apartments of the Habitat Matterport 3D Semantics dataset~\cite{habitat23semantics}. Yellow nodes represent floors, while blue nodes represent rooms. The green graph right above the respective floor represents the produced navigational graph. For more visualizations, please refer to Fig.~\ref{fig:hm3dsem-sg-vis-suppl}.}} 
\label{fig:hm3dsem-sg-vis}
\end{figure*}

\subsubsection{Semantic Room Classification}
\label{sec:exp-semantic-room-classification}
\color{black}
We quantitatively evaluate our proposed view embedding-based room category labeling method (See Fig.~\ref{fig:room_embedding}) by comparing it against two strong baselines across the set of eight scenes on HM3DSem. Both baselines rely on object labels to classify room categories. To draw a fair comparison, all methods rely on ground truth room segmentation, namely the class-agnostic mask of each room. Thus, all objects are assigned to rooms based on ground truth room layouts. This is different from the general \ours{} method, which also estimates room masks.

\noindent\textbf{Dataset:} In this evaluation, we utilize a closed set of room categories. The HM3DSem dataset does not provide annotated room categories but merely educated votes, which are often not sufficient. Therefore, we manually labeled the regions of the eight scenes detailed in Sec.~\ref{sec:exp_graph_eval_hm3dsem}. The list of room categories is provided in Sec.~\ref{sec:room-classification-suppl}. 

\noindent\textbf{Baselines:} We compare the \ours{} approach of using filtered view embeddings for labeling rooms against a privileged and an unprivileged baseline. The privileged baseline operates on ground truth object categories contained within each room. In order to obtain room labels, the baseline prompts an LLM (GPT-3.5 and GPT-4~\cite{openai2023gpt}) to provide a room category guess out of the closed set of room categories given the objects per room in a few-shot manner (prompts are detailed in Sec.~\ref{sec:room-classification-suppl}). The second and unprivileged baseline applies the same principle of prompting an LLM but only has access to the predicted object categories obtained using \ours{}. This means that each predicted object is labeled as the category showing the highest similarity to the object feature among a category list. In general, we expect that the number of objects will be different from the privileged baseline because of under- and over-segmentation of \ours{}'s predictions. In comparison, our view embedding method relies on 10 distinct view embeddings, which are scored against the chosen set of room categories. The final predicted room category is defined by the room category that showed the highest similarity across all view embeddings as described in Fig.~\ref{fig:room_embedding}.

\noindent\textbf{Metrics:} We apply two different evaluation criteria: The first accuracy called $\text{Acc}_{=}$ fosters replicability by evaluating whether the predicted and the ground truth room categories are text-wise equal. Different from that, the performance regarding the $\text{Acc}_{\approx}$ metric is produced via human evaluation. This is crucial as room categories are not always fully determinable when labeling, e.g., combined kitchen and living room areas. Moreover, the answers provided by the LLM do not always state definitive categories because of frequent hallucinations. A high number of objects per room exacerbates this. This particularly applies to the unprivileged baselines when facing under-segmentation. In order to circumvent this, we manually filter all outputs across the set of eight scenes and check whether the LLM \textit{leaned} towards the correct answer such as predicting a synonym of the GT room type, which boosts results in favor of the LLM-based methods. In addition to this, we also evaluated the same task using the current state-of-the-art LLM, GPT-4, which shows significantly fewer hallucinations and increased accuracy.

\noindent\textbf{Results:} As presented in Table~\ref{tab:habitat-room-eval}, the view embedding method of \ours{} outperforms all unprivileged baselines both in terms of the strict accuracy ($\text{Acc}_{=}$) as well as the approximate accuracy ($\text{Acc}_{\approx}$) by a significant margin. In addition, \ours{} also outperforms the privileged baseline relying on GPT-3.5 while showing similar approximate accuracy as GPT-4 of 84.10\% vs.\ 84.25\%.
Thus, we conclude that our room labeling method is robust and outperforms comparable unprivileged methods by a significant margin. Furthermore, we provide additional scene-wise evaluations in Table~\ref{tab:habitat-room-eval-suppl} in the supplementary material.
\color{black}

\begin{table}[h]
\color{black}
\centering
\footnotesize
\caption{\textsc{Semantic Room Classification Results (HM3DSem)}}
\setlength\tabcolsep{13pt}
\begin{threeparttable}
\begin{tabular}{cl|cc}
 \toprule
\multicolumn{2}{l}{Room Identification Method} & $\text{Acc}_{=}$\,[\%] & $\text{Acc}_{\approx}$\,[\%] \\
\midrule 
\multirow{5}{*}{\rotatebox[origin=c]{90}{Privileged}} &
GPT-3.5 w/ GT & \multirow{2}{*}{66.89} & \multirow{2}{*}{81.49} \\
& object categories &  &  \\
\cmidrule{2-4}
& GPT-4 w/ GT & \multirow{2}{*}{79.86} & \multirow{2}{*}{84.25} \\
& object categories & & \\
\midrule
\midrule 
\multirow{7}{*}{\rotatebox[origin=c]{90}{Unprivileged}} &
GPT-3.5 w/ predicted & \multirow{2}{*}{28.48} & \multirow{2}{*}{42.95} \\
& object categories & &  \\
\cmidrule{2-4}
& GPT-4 w/ predicted & \multirow{2}{*}{\underline{59.47}} & \multirow{2}{*}{\underline{62.55}} \\
& object categories &  & \\
\cmidrule{2-4}
\cmidrule{2-4}
& HOV-SG (ours) & \multirow{2}{*}{\textbf{73.93}} & \multirow{2}{*}{\textbf{84.10}} \\
& w/ view embeddings &  & \\
\bottomrule
\end{tabular}
\footnotesize
We present the room classification performance on HM3DSem of \ours{} (view embeddings) and two baselines (GPT-3.5 / GPT-4) using either privileged or unprivileged information. In the privileged case, rooms are labeled based on ground truth object categories per room. The unprivileged baselines take the predicted masks and categories as input. We consider two different evaluation criteria: $\text{Acc}_{\text{=}}$ measures whether the exact text-wise room category was predicted while $\text{Acc}_{\approx}$ measures semantically correct room predictions based on qualitative human evaluation through manual inspection.
\end{threeparttable}
\label{tab:habitat-room-eval}
\end{table}

\begin{figure*}
\centering
\footnotesize
\includegraphics[width=0.9\textwidth]{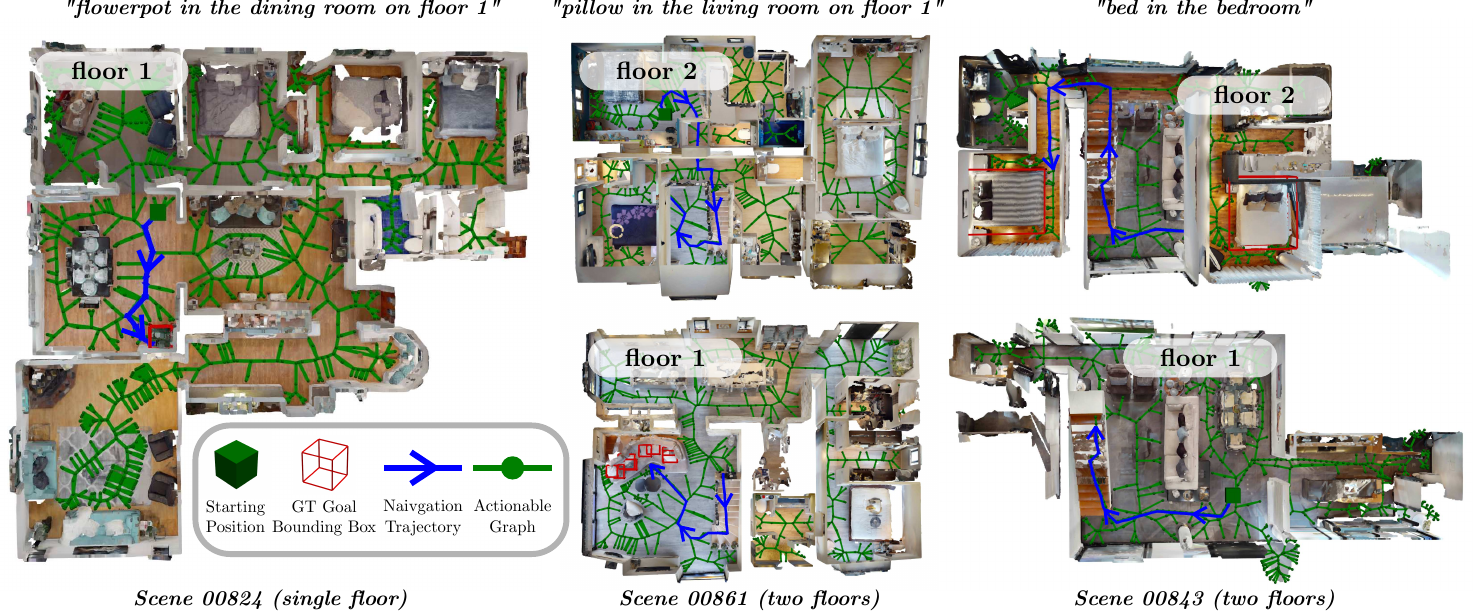}
\caption{\rebuttal{Qualitative visualization of the language-conditioned navigation in multi-floor environments in HM3DSem dataset. In some cases, there are multiple ground truth objects matching the instruction. Stopping at any of them with a distance within 1 meter is regarded as a success.}} 
\label{fig:lang_nav_traj}
\end{figure*}

\subsubsection{Object-Level Semantics}
\label{sec:exp-hm3dsem-objects}
\rebuttal{Existing open-vocabulary evaluations usually circumvent the problem of measuring true open-vocabulary semantic accuracy. This is due to arbitrary sizes of the investigated label sets, a potentially enormous and challenging amount of object categories~\cite{habitat23semantics}, and the ease of use of existing evaluation protocols~\cite{huang23vlmaps, conceptfusion}. While human-level evaluations such as Amazon Mechanical Turk (AMT) partly solve this problem, robust results replication and scaling remain challenging~\cite{conceptgraphs}.}

\noindent\rebuttal{\textbf{Metrics:} In this work, we propose the novel AUC$_{k}^{top}$ metric that quantifies the area under the top-$k$ accuracy curve between the predicted and the actual ground-truth object category (see Fig.~\ref{fig:auc-vis}). This entails computing the ranking of all cosine similarities between the predicted object feature and all possible category text features, which are in turn encoded using a vision-language model (CLIP). Thus, the metric encodes how many \textit{erroneous shots} are necessary on average before the ground-truth label is predicted correctly. Based on this, the metric encodes the actual open-set similarity while scaling to large, variably-sized label sets. We envision a future use of this metric in diverse open-vocabulary tasks.}

\rebuttal{We visualize the AUC$_{k}^{top}$ curve \ours{} on the \textit{00824} scene of HM3DSem in Fig.~\ref{fig:auc-vis}. The closer the curve is to the upper left corner of the plot, the higher the open-vocabulary similarity. Instead of showing the accuracy at distinct values of $k$, we normalize $k$ over the extent of the label set, which contains 1624 categories for HM3DSem. This also shows visually how the AUC$_{k}^{top}$ metric provides a dependable measure for large but variably sized label sets.}

\noindent\rebuttal{\textbf{Baselines:} In order to show the applicability of the AUC$_{k}^{top}$ metric, we compare \ours{} against two strong baselines VLMaps~\cite{huang23vlmaps} and ConceptGraphs~\cite{conceptgraphs} on the Habitat-Semantics dataset~\cite{habitat23semantics}, which comprises an enormous label set of 1624 object categories. To allow for a fair comparison, we perform a linear assignment among predicted and GT objects and only consider predicted objects that show an $\operatorname{IoU}>50\%$ with the ground truth. Since VLMaps~\cite{huang23vlmaps} does not predict masks by design, it takes the object masks predicted by \ours{} and averages all masked voxels' features as the object feature.}

\noindent\rebuttal{\textbf{Results:} The overall AUC$_{k}^{top}$ score as well as various top-k thresholds are given in Table~\ref{tab:seg-auc}. We observe that VLMaps~\cite{huang23vlmaps} performs inferior, which is presumably due to its dense feature aggregation as well as its dependence on a fine-tuned VLM, LSeg~\cite{li2022language}, limiting its generalization in challenging open-vocabulary scenarios. It does not only score merely 5\% of objects correctly given its top-5 choices but also only predicts the correct class given its top-500 predicted classes in 40.02\% of all cases. In comparison, ConceptGraphs~\cite{conceptgraphs} obtains a competitive score of 84.07\% while \ours{} achieves 84.88\%. Especially, up to the top-100 highest ranking classes, \ours{} outperforms ConceptGraphs.}

\vspace{0.3cm}

\begin{table}[t]
\centering
\scriptsize
\caption{\textsc{Object-Level Semantics Evaluation on HM3DSem}}
\setlength\tabcolsep{3pt}
\begin{threeparttable}
\begin{tabular}{l|cccccc|c}
 \toprule

Method  & $top_{5}$ & $top_{10}$ & $top_{25}$ & $top_{100}$ & $top_{250}$ & $top_{500}$ & AUC$_{k}^{top}$ \\
\midrule
 VLMaps \cite{huang23vlmaps} & 0.05 & 0.17 & 0.54 & 15.32 & 26.01 & 40.02 & 56.20  \\

ConceptGraphs~\cite{conceptgraphs} & 18.11 & 24.01 & 33.00 & 55.17 & \textbf{70.85} & \textbf{81.55}  & \underline{84.07} \\
 \midrule
HOV-SG (ours) & \textbf{18.43} & \textbf{25.73} & \textbf{36.41} & \textbf{56.46} & 69.95 & 80.86 & \textbf{84.88} \\

\bottomrule
\end{tabular}

\footnotesize
We provide object-level semantic accuracies across all eight considered scenes within HM3DSem using both the overall AUC$_{k}^{top}$ metric across 1624 categories as well as accuracies at a few selected thresholds $k$. To allow for a fair comparison, we perform a linear assignment among predicted and GT objects and only consider predicted objects that show an $\operatorname{IoU}>50\%$ with the ground truth. Since VLMaps~\cite{huang23vlmaps} does not predict masks by design, it takes the masks predicted by HOV-SG and evaluates wrt.\ those.
\end{threeparttable}
\label{tab:seg-auc}
\end{table}

\subsubsection{Hierarchical Concept Retrieval}
\label{sec:exp-hier-concept-retrieval}
\rebuttal{To take advantage of the hierarchical character of our proposed representation, we evaluate to what extent we can retrieve objects from hierarchical queries of the form: \textit{``pillow in the living room on the second floor''} or \textit{``bottle in the kitchen''}. To do so, we decompose the query using GPT-3.5 into its sought-after hierarchical concepts, e.g., \texttt{[floor 2}, \texttt{living room}, \texttt{pillow]} or \texttt{[-}, \texttt{kitchen}, \texttt{bottle]}), and compute the corresponding CLIP embeddings. In the next step, we hierarchically query against the most suitable floor, the most appropriate room, and lastly, the most suitable object given the query at hand (see Table~\ref{tab:habitat-retrieval}). While floor prompting is done naively, we select the room producing the highest maximum cosine similarity to the query room across its ten embeddings. On average, this produces higher success rates compared with mean- or median-based schemes.}

\noindent\rebuttal{\textbf{Retrieval Results:} In the following, we compare \ours{} against an augmented variant of ConceptGraphs~\cite{conceptgraphs} that is equipped with privileged floor information and it scores objects against the requested room and object, which allows it to draw answers at the floor and room level. As shown in Table~\ref{tab:habitat-retrieval}, \ours{} shows a significant performance increase of 11.69\% on object-room-floor queries and a 2.2\% advantage on object-room queries when compared with ConceptGraphs. While ConceptGraphs struggles on larger scenes and under more detailed queries, \ours{} outperforms it by a significant margin even though it suffers from erroneous room segmentations by design. For more information, we refer to Sec.~\ref{subsec:prompt-supp}.}

\begin{table}[h]
\color{black}
\centering
\scriptsize
\caption{\textsc{Object Retrieval from Language Queries (HM3DSem)}}
\setlength\tabcolsep{3.7pt}
\begin{threeparttable}
\begin{tabular}{ll|ccccc}
 \toprule
Query Type & Method & \#\,Trials & Retrieval-SR$_{10}$\,[\%] & Navigation-SR\,[\%]\\
\midrule
\multirow{2}{*}{(\texttt{o}, \texttt{r}, \texttt{f})} & ConceptGraphs & \multirow{2}{*}{40.63} & 16.31 & - \\
& {HOV-SG (ours)} &  & \textbf{28.00} & 37.32 \\
 \midrule
\multirow{2}{*}{(\texttt{o}, \texttt{r})} & ConceptGraphs & \multirow{2}{*}{34.88} & 29.26 & - \\
& {HOV-SG (ours)} &  & \textbf{31.48} & 40.41 \\
\bottomrule
\end{tabular}
\footnotesize
Evaluation over 20 frequent distinct object categories in terms of the top-5 accuracy. A match is counted as a success when the $\text{IoU}>0.1$ between predicted object and ground truth. The number of trials is an average number of trials across the eight scenes evaluated. It is lower for \texttt{(o,r)} compared to \texttt{(o,r,f)} due to a higher number of query duplicates whenever we drop the floor specification. The 20 evaluated categories are: \textit{picture, pillow, door, lamp, cabinet, book, chair, table, towel, plant, sink, stairs, bed, toilet, tv, desk, couch, flowerpot, nightstand, faucet}.
\end{threeparttable}
\label{tab:habitat-retrieval}
\color{black}
\end{table}

\noindent\rebuttal{
\textbf{Language-Grounded Navigation in Simulation:} In addition, to the general querying tasks we also perform navigational trials using the Habitat Simulator. Based on the obtained actionable multi-floor navigational Voronoi graphs introduced in Sec.~\ref{subsec:scene_graph_creation}, we traverse the set of retrieved high-probability objects satisfying the query and report the physical retrieval success rate. If the robot stops by one of the matched ground truth point clouds and its distance to it is smaller than \SI{1}{m}, we consider the navigation successful. As shown in Table~\ref{tab:habitat-retrieval}, the navigational success rates are higher when compared to the retrieval success rates. We found that the robot regularly reaches the locations of the predicted objects that are in close vicinity to actual ground truth target objects. This effect is measurable since we rely on a Euclidean distance-based evaluation for the navigational success assessment. Moreover, imperfectly predicted instance masks increase the chance of retrieving segments that do not provide complete overlap with actual ground truth objects in terms of geometry and semantics, which induces slight offsets in the retrieved positions when querying. Example navigation trials are shown in Fig.~\ref{fig:lang_nav_traj}.
}

\begin{table}
\centering
\footnotesize
\caption{\centering\textsc{Real-World Object Retrieval and Goal Navigation from Language Queries}}
\setlength\tabcolsep{8pt}
\begin{threeparttable}
\begin{tabular}{lc|cc|ccc}
 \toprule
Query & \# & \multicolumn{2}{c|}{\text{Graph Querying}} & \multicolumn{3}{c}{\text{Goal Navigation}} \\
Type & \text{Trials} & \text{\# Successes} & \text{SR\,[\%]}  & \text{Success} & \text{SR\,[\%]}  \\
\midrule
\text{Object} & 41 & 29 & 70.7 & 23 & 56.1  \\
\text{Room} & 9 & 5 & 55.6 & 5 & 55.6 \\
\text{Floor} & 2 & 2 & 100 & 2 & 100 \\
\bottomrule
\end{tabular}
\footnotesize
We count a retrieval as successful whenever the robot is in close vicinity to the object sought after ($\sim\SI{1}{m}$).
\end{threeparttable}
\label{tab:real-world-retrievel}
\end{table}

\subsection{Real World Language-Grounded Navigation}
\label{sec:exp_lang_nav_real_world}
To validate the system in the real world, we utilize a Boston Dynamics \textit{Spot} robot with a calibrated Azure Kinect RGB-D camera and a 3D LiDAR attached to it. We collect a stream of RGB-D sequences inside a \rebuttal{two-story} office building, traversing through a variety of rooms with diverse semantic information as shown in Fig.~\ref{fig:spot_overview}. 

\begin{figure}[t]
\includegraphics[width=\columnwidth]{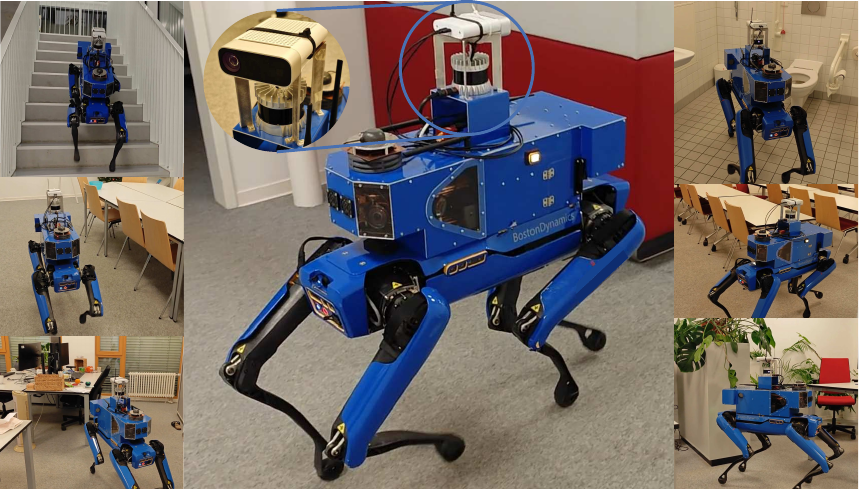}
\caption{Boston Dynamics \textit{Spot} robot exploring a \rebuttal{two-story} office building with multiple types of rooms. The quadruped is equipped with an Azure Kinect RGB-D camera and a 3D LiDAR to collect RGB-D data and odometry.}
\label{fig:spot_overview}
\vspace{-2em}
\end{figure}

\noindent\textbf{Real-World Application of HOV-SG}: We calibrate the extrinsics between the LiDAR and the RGB-D camera and \rebuttal{apply an off-the-shelf LiDAR SLAM implementation that combines \rebuttal{FAST-LIO2}~\cite{xu2022fast} with a loop closure component to obtain LiDAR poses}. Subsequently, we leverage the extrinsics to derive the associated camera poses. Finally, employing the RGB-D data and odometry, we construct the \ours{} representation as detailed in Sec.~\ref{sec:approach}.

\noindent\textbf{Robot Navigation with Long Queries}: Within the two-story building, we select 41 object goals, nine room goals, and two floor goals and use natural language to query the HOV-SG representation built. Some examples of the queries are ``go to floor 0'', ``navigate to the kitchen on floor 1'', and ``find the plant in the office on floor 0''. Unlike previous methods that only retrieved object-level goals, our representation enables long queries containing multiple levels of concepts and facilitates a more detailed constrained retrieval. 

To separate the evaluation of our representation and the navigation system, we first evaluate the accuracy of the retrieval tasks qualitatively. Since the building is well structured, we can easily determine the boundary between rooms and regions like offices, corridors, and dining rooms. Meanwhile, we manually label the categories of all regions in the building, using a category set containing \texttt{office, corridor, kitchen, seminar room, meeting room, dining room, bathroom}. If the HOV-SG representation returns the correct floor and room point cloud as well as an object point cloud that shows overlap with the correct object, we consider this as retrieval success. We achieve a 100\% success rate in floor retrieval, a 55.6\% success rate in room retrieval, and a 70.7\% success rate in object retrieval. The major failure cases for room retrieval stem from the visual similarity among rooms such as ``meeting room'', ``seminar room'', and ``dining room'' as shown in Fig.~\ref{fig:room_comparison}.


We query our HOV-SG representation using hierarchical concepts and utilize the \textit{Spot} quadrupedal robot to carry out navigation trials. In our experiments, the robot navigates to queried objects across both floors with a 56.1\% success rate while navigating to all successfully retrieved room and floor concepts with language instructions. One failure case occurred for the ``go to the whiteboard in the office on the second floor'' query. Since the whiteboard is attached to a room-separating wall, the robot achieved the necessary distance to the object but was positioned on the opposite side of the wall as shown in Fig.~\ref{fig:real_world_target_object}). In addition to that, we did not consider target locations on stair segments to prevent the robot assuming unstable poses. The evaluation results for retrieval and navigation are shown in Table~\ref{tab:real-world-retrievel}. We display a subset of target objects in Fig.~\ref{fig:real_world_target_object} and three trials in more detail in the supplementary material Fig.~\ref{fig:real-world-trials}.

\subsection{Representation Storage Overhead Evaluation}
\label{sec:exp_size}
\rebuttal{A key advantage of \ours{} is the compactness of the representation. We compare the storage size of VLMaps~\cite{huang23vlmaps}, ConceptGraphs~\cite{conceptgraphs}, and \ours{} created for the eight scenes in the HM3DSem dataset and show the results in Tab.~\ref{tab:size}. We adapt VLMaps to store LSeg features at 3D voxel locations with 0.05m voxel size. The backbone of the LSeg is ViT-B-32, which has 512-dimensional features. ConceptGraphs and \ours{} are using the ViT-H-14 CLIP backbones, which requires saving 1024-dimension features in the representation. VLMaps is optimized to only save features at voxels near object surfaces instead of saving redundant features at non-occupied voxels. Nonetheless, thanks to the compact graph structure, ConceptGraphs and \ours{} are much smaller than their dense counterparts. \ours{} even reduces as much as 75\% in memory footprint on average compared to VLMaps. While ConceptGraphs encodes supplementary object relationships, \ours{} incorporates hierarchical semantic features. Overall, ConceptGraphs and \ours{} serve as excellent complementary representations, each emphasizing distinct facets of scene semantics.}


\begin{table}
\centering
\scriptsize
\caption{\textsc{Representation Size (HM3DSem)}}
\setlength\tabcolsep{4pt}
\begin{threeparttable}
\begin{tabular}{ccccc}
\toprule
Scene & \# Floors & VLMaps~\cite{huang23vlmaps} & ConceptGraphs~\cite{conceptgraphs} & HOV-SG (ours) \\ 
\midrule
00824 & 1 & 568    & \textbf{143}           & \textbf{143}  \\
00829 & 1 & 407    & 110           & \textbf{99}   \\
00843 & 2 & 534    & 143           & \textbf{125}  \\
00861 & 2 & 943    & 255  & \textbf{225}  \\
00862 & 3 & 1808   & \textbf{474}  & 479  \\
00873 & 2 & 570    & 167           & \textbf{129}  \\
00877 & 2 & 556    & 154           & \textbf{131}  \\
00890 & 2 & 682    & 192           & \textbf{162}  \\
\midrule
$\Sigma$   & - & 6068   & 1638          & \textbf{1493} \\
\bottomrule
\end{tabular}
\footnotesize
We compare the storage sizes of the representations produced by VLMaps~\cite{huang23vlmaps}, ConceptGraphs~\cite{conceptgraphs}, and \ours, measured in megabyte (MB), across eight differently sized scenes of Habitat-Semantics (HM3DSem). The smallest sizes are highlighted \textbf{bold}, respectively.
\end{threeparttable}
\label{tab:size}
\vspace{-2em}
\end{table}

\subsection{Ablation Study}
\label{sec:exp_ablation}
\rebuttal{In order to shed light on the contributions of various key components in our approach, we present an ablation study on the Replica dataset~\cite{straub2019replica} in Table~\ref{tab:seg-ablation}. One key component of the 3D open-vocabulary segment-level mapping pipeline (Sec.~\ref{subsec:segment_map_creation}) is the DBSCAN clustering we apply to pixel-wise CLIP embeddings associated with each segment to select the most representative features among them. This design, inspired by the principle of majority voting, has proven effective in mitigating outlier CLIP features caused by the inherent limitations of CLIP and the noise originating from SAM's outputs, thereby enhancing semantic accuracy. A different key component of our approach involves fusing CLIP features extracted from various sources: the global image, the masked image of an object based on its SAM mask, and the masked object image given the SAM mask without background. In contrast to ConceptGraphs~\cite{conceptgraphs}, which only integrate the global image embedding and the sub-image embedding, we hypothesize that incorporating salient features from the sub-image into CLIP embeddings could enhance accuracy. Based on this, we tested three setups: utilizing only the CLIP embedding of the masked object including background (L-CLIP), employing only the CLIP embedding of the masked object without background (M-CLIP), and third, combining both of these CLIP embeddings by fusing them as done in \ours{}. Our findings indicate that combining both embeddings yields the highest semantic accuracy as given in Table~\ref{tab:seg-ablation}.}


\begin{table}[t]
\color{black}
\centering
\begin{threeparttable}

\scriptsize
\setlength\tabcolsep{4pt}
\caption{\textsc{Ablation Study on Replica}}
\begin{tabular}{ccc|ccc}
\toprule
DBSCAN  & L-CLIP & M-CLIP & mIOU$\,\uparrow$ & F-mIoU$\,\uparrow$ & mAcc$\,\uparrow$ \\
\midrule
\xmark & \cmark & \cmark & 0.212 & 0.340 & 0.290 \\
\cmark & \xmark & \cmark & 0.136 & 0.178 & 0.170  \\
\cmark & \cmark & \xmark & 0.215 & 0.337 & 0.298  \\
\midrule
\multicolumn{3}{c|}{\ours{} (ours)} & \textbf{0.231} & \textbf{0.386} & \textbf{0.304}  \\
\bottomrule
\end{tabular}
\footnotesize
DBSCAN indicates whether we apply DBSCAN clustering to select segment features, L-CLIP indicates the use of only masked images including background, and M-CLIP refers to only the masked CLIP embeddings without background.
\label{tab:seg-ablation}
\end{threeparttable}
\color{black}
\end{table} 
\section{Conclusion}
\label{sec:conclusion}

\rebuttal{We presented \ours{}, a novel hierarchical open-vocabulary 3D scene graph representation for indoor robot navigation. Through the semantic decomposition of environments into floors, rooms, and objects, we demonstrate effective concept retrieval from abstract language queries and perform long-horizon navigation across a multi-story environment in the real world. With extensive experiments conducted across multiple datasets, we showcase that \ours{} surpasses previous baselines in terms of semantic accuracy, open-vocabulary capability, and compactness. Nevertheless, \ours{} is not without limitations. Consisting of several stages and components, our approach necessitates a large number of hyper-parameters. Moreover, the construction process of \ours{} is time-consuming, rendering the method unsuitable for real-time mapping. Furthermore, it assumes a static environment and thus cannot handle dynamic environments. Future research directions may involve developing an open-vocabulary dynamic representation of the environment or integrating a reactive embodied agent to enhance reasoning and grounding in the physical world. To foster future research, we make the code publicly available at \website{}.}

\vspace{1em}
\section{Acknowledgement}
This work was funded by the German Research Foundation (DFG) Emmy Noether Program grant number 468878300, the BrainLinks-BrainTools Center of the University of Freiburg, and an academic grant from NVIDIA.


\bibliographystyle{ieeetr}
\bibliography{references}

\clearpage

\maketitle

\normalsize
\clearpage
\renewcommand{\baselinestretch}{1}
\setlength{\belowcaptionskip}{0pt}

\begin{strip}
\begin{center}
\vspace{-5ex}
\textbf{\LARGE \bf
Hierarchical Open-Vocabulary 3D Scene Graphs \\\vspace{0.5ex} for Language-Grounded Robot Navigation
} \\
\vspace{3ex}

\Large{\bf- Supplementary Material -}\\
\vspace{0.6cm}
\normalsize{Abdelrhman Werby\textsuperscript{\footnotesize 1*},\quad
Chenguang Huang\textsuperscript{\footnotesize 1*},\quad
Martin Büchner\textsuperscript{\footnotesize 1*},\quad
Abhinav Valada\textsuperscript{\footnotesize 1}, \quad
Wolfram Burgard\textsuperscript{\footnotesize 2}}\\
\vspace{0.3cm}
\normalsize{\textsuperscript{\footnotesize 1}University of Freiburg \quad\quad {\textsuperscript{\footnotesize 2}University of Technology Nuremberg}}

\end{center}
\end{strip}

\setcounter{section}{0}
\setcounter{equation}{0}
\setcounter{figure}{0}
\setcounter{table}{0}
\setcounter{page}{1}
\makeatletter

\renewcommand{\thesection}{S.\arabic{section}}
\renewcommand{\thesubsection}{S.\arabic{section}-\Alph{subsection}}
\renewcommand{\thetable}{S.\arabic{table}}
\renewcommand{\thefigure}{S.\arabic{figure}}

In this supplementary material, we expand upon multiple
aspects of our main paper. In Sec.~\ref{sec:suppl_method}, we detail several design choices and paradigms regarding our method.
In Sec.~\ref{sec:suppl_experiments}, we additionally present experimental results that support the claims introduced in the manuscript. This includes a more detailed discussion of the proposed open-vocabulary metric, an analysis regarding identified semantic room categories, additional baselines regarding object retrieval from language queries.
Moreover, we provide insightful visualizations of the produced multi-story scene graphs, representing scenes from both Habitat Semantics (HM3DSem) as well as our real-world environment. 

\section{Method}
\label{sec:suppl_method}
\begin{figure}[t]
\includegraphics[width=0.5\textwidth]{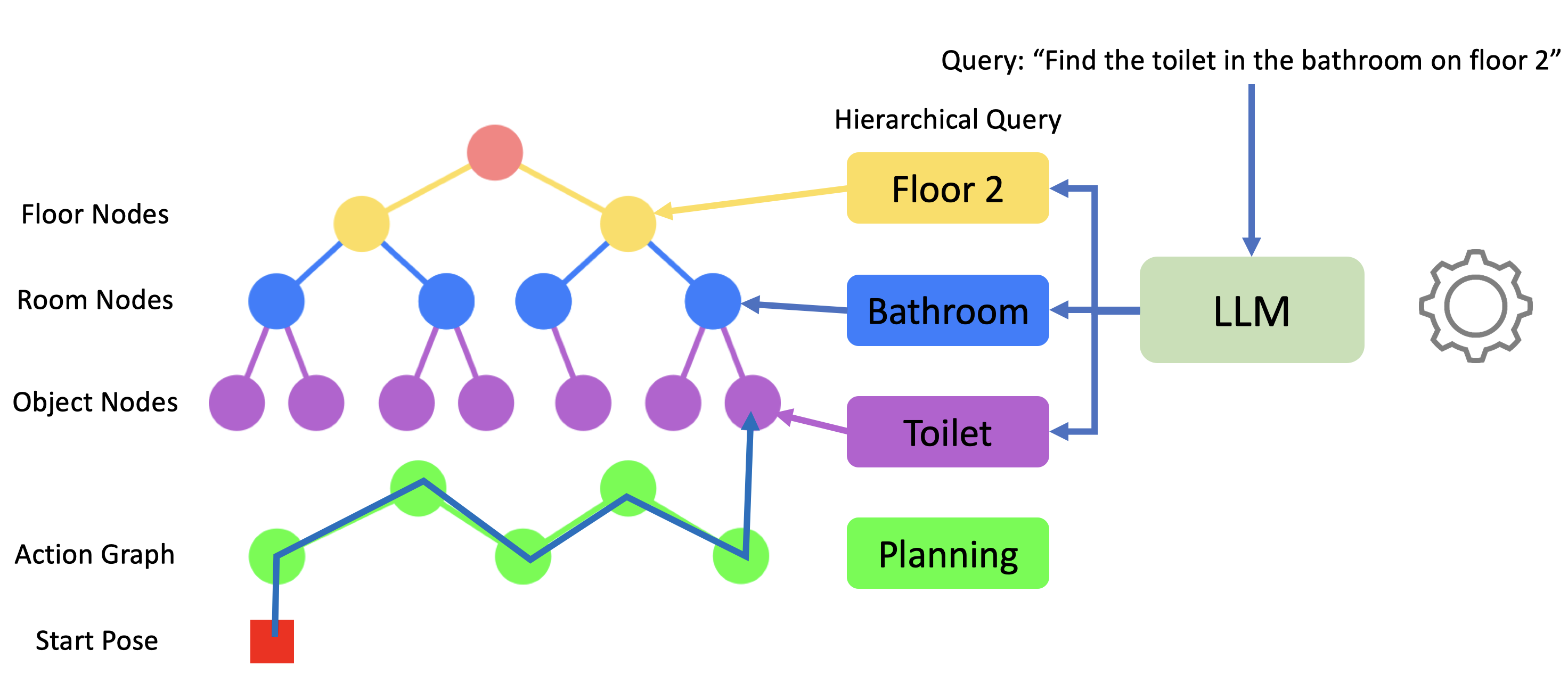}
\caption{The language-grounded navigation module of HOV-SG allows to parse complex queries such as ``find the toilet in the bathroom on floor 2'' into three queries using a large language model (GPT-3.5) - one each for the floor, room, and object levels. Leveraging HOV-SG's hierarchical structure, we progressively narrow down the search space by querying at each level. Once the target location is identified, the action graph in HOV-SG is used to plan a path from the starting pose to the target.}
\label{fig:long_query_nav_approach}
\end{figure}

\subsection{Robot Navigation Prompts}
\label{subsec:prompt-supp}
\rebuttal{We provide few-shot examples to GPT-3.5 or GPT-4 to increase the chance of correctly decomposing the instruction query at test time. In the following, we detail the ``system'', ``user'', and ``assistant'' variables used in {\color{prompt-gray}gray}, \command{green}, and \colorbox{highlight}{highlighted}, respectively. ``INSTRUCTION'' is the test query provided by the user. The prompt is shown below:}
\lmp{
    \scriptsize
    \prompt{system: You are a hierarchical concept parser. You need to parse a description of an object into floor, region and object.}\\
    \command{Q: chair in region living room on the 0 floor}\\
    \colorbox{highlight}{A: [floor 0,living room,chair]}\\
    \command{Q: floor in living room on floor 0}\\
    \colorbox{highlight}{A: [floor 0,living room,floor]}\\
    \command{Q: table in kitchen on floor 3}\\
    \colorbox{highlight}{A: [floor 3,kitchen,table]}\\
    \command{Q: cabinet in region bedroom on floor 1}\\
    \colorbox{highlight}{A: [floor 1,bedroom,cabinet]}\\
    \command{Q: bedroom on floor 1}\\
    \colorbox{highlight}{A: [floor 1,bedroom,]}\\
    \command{Q: bed}\\
    \colorbox{highlight}{A: [,,bed]}\\
    \command{Q: bedroom}\\
    \colorbox{highlight}{A: [,bedroom,]}\\
    \command{Q: I want to go to bed, where should I go?}\\
    \colorbox{highlight}{A: [,bedroom,]}\\
    \command{Q: I want to go for something to eat upstairs. I am currently at floor 0, where should I go?}\\
    \colorbox{highlight}{A: [floor 1,dinning,]}\\
    \command{Q: INSTRUCTION} \\
}


\subsection{Semantic Localization}
HOV-SG achieves agent localization within the graph using only RGB images and local odometry using a simple particle filter. The process involves randomly initializing K particles within the free space map, estimated from each floor's point cloud. Subsequently, the global CLIP feature of the RGB image and the CLIP feature of objects within the image are extracted using the same pipeline as used for the graph creation. In the prediction step, the particle poses are updated based on robot odometry. Thus, we assign each particle a floor and room based on its updated coordinates. In the update step, we calculate cosine similarity scores between the current RGB image's global CLIP feature and the graph's room features for each particle. Additionally, scores are computed between object features in the RGB image and observed objects in front of each particle. Then, particle weights are adjusted based on these similarity scores. This integrated approach allows HOV-SG to semantically localize the agent within the graph at the floor and room level within a short span of 10 observed frames.

\label{sec:suppl_sem_loc}

\section{Experimental Evaluation}
\label{sec:suppl_experiments}

\subsection{Open-Vocabulary Similarity Metric (AUC$_{k}^{top}$)}
In this section, we present a visualization of the open-vocabulary similarity metric AUC$_{k}^{top}$ introduced in the paper. As shown in Fig.~\ref{fig:auc-vis}, the AUC$_{k}^{top}$ metric represents the area under the top-k accuracy curve. The closer this curve is to the upper left point, the higher the open-vocabulary similarity. Instead of showing the accuracy at distinct values of $k$ as in the main paper, we normalize $k$ over the extent of the label category set, which contains 1624 categories for HM3DSem. This also shows visually how the AUC$_{k}^{top}$ metric provides a dependable measure for large but variably-sized label sets. We envision the future use of this metric in a number of open-vocabulary tasks.

\begin{figure}
\centering
\footnotesize
\begin{tikzpicture}
\begin{axis}[xlabel=\% of ranked categories considered,
ylabel=Accuracy, 
grid=both,
width=7cm,height=5cm,
legend cell align=left,
legend pos=south east,
ymax=100, ymin=0,
xmin=-5, xmax=100]
  \addplot[name path=hov, color=red,mark=*] coordinates {
		(0.0, 0)
            (0.3,21.38)
            (1.54, 40.13)
            (6.16, 64.14)
            (15.39, 75.66)
		(30.78, 82.22)
            (61.57, 90.32)
		(100, 100)
	};
  \addplot+[draw=none,name path=B, mark=none] coordinates {
            (0,0) 
            (100,0)
        };
 \node[text=red] at (axis cs: 50,50) {AUC$_{k}^{top}$ = 86.52 };
 \legend{HOV-SG (ours)};
 \addplot+[red!10, opacity=0.4] fill between[of=hov and B,soft clip={domain=0:100}]; 
\end{axis}
\end{tikzpicture}
\caption{We visualize the AUC$_{k}^{top}$ curve for different evaluation thresholds $k$, which this plot measures in terms of percent out of the total number of categories (HM3DSem: 1624). The shown curve represents the results of our method HOV-SG on the HM3DSem scene \textit{00824}.}
\label{fig:auc-vis}
\end{figure}
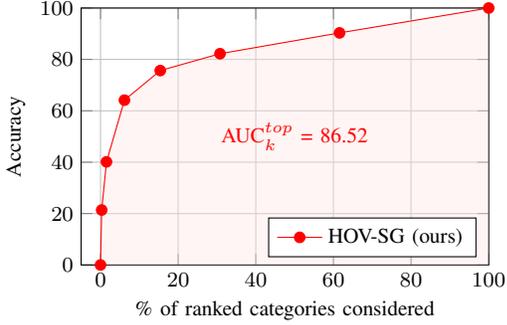

\subsection{Evaluation Metrics for 3D Semantic Segmentation}
\label{sec:evaluation-metrics-3d-seg-suppl}
\rebuttal{In terms of 3D semantic segmentation, we perform point-wise evaluation between the predicted and ground truth labels. The used 3D semantic segmentation metrics mIOU, F-mIOU, and mAcc are defined as follows:} 

\rebuttal{
\begin{equation}
    \text{mIOU} = \frac{1}{N} \sum^{N}_{i=1}\frac{TP_i}{TP_i+FP_i+FN_i}
\end{equation}
}
\rebuttal{
\begin{equation}
    \text{F-IOU} = \frac{1}{\sum^{N}_{i=1}n_i} \sum^{N}_{i=1}\frac{n_i \cdot TP_i}{TP_i+FP_i+FN_i}
\end{equation}
}
\rebuttal{
\begin{equation}
    \text{mAcc} = \frac{1}{N} \sum^{N}_{i=1}\frac{TP_i}{TP_i+FP_i}
\end{equation}
}
\rebuttal{
where $N$ is the total number of categories, $TP_i$, $FP_i$, and $FN_i$ denote the truth positive, false positive, and false negative number for category $i$, and $n_i$ is the total number of ground truth points for category $i$.
}

\subsection{Floor and Region Segmentation}
\label{sec:floor-room-seg-suppl}
\color{black}
In addition to the overall results provided in Table~\ref{tab:floor-region}, we provide scene-wise results in Table~\ref{tab:floor-region-suppl}. Our results show that we tend to obtain higher precision and recalls on smaller scenes comprising fewer ground-truth rooms.
\color{black}
\begin{table}[t]
\color{black}
\centering
\scriptsize
\caption{\textsc{Floor and Region Segmentation on HM3DSem}}
\setlength\tabcolsep{5.7pt}
\begin{threeparttable}
\begin{tabular}{cc|ccc|cc}
 \toprule
\multirow{2}{*}{Method} & \multirow{2}{*}{Scene}  & \multicolumn{3}{c|}{Floors} & \multicolumn{2}{c}{Regions} \\
&  & $\text{Acc}_{F}$ & $\text{\#\,F}_{\text{P}}$ & $\text{\#\,F}_\text{GT}$ &  
 $\text{Precision}$\,[\%] &  $\text{Recall}$\,[\%] \\
 \midrule 
\multirow{8}{*}{\rotatebox[origin=c]{90}{Hydra \cite{hughes2022hydra}}} & \textit{00824} & -& - & - & 79.89 & 78.37 \\
& \textit{00829} & - & - & - & 86.42 & 85.08 \\ 
& \textit{00843} & - & - & - & 87.27 & 78.77 \\ 
& \textit{00861} & - & - & - & 77.23 & 67.93 \\ 
& \textit{00862} & - & - & - & 85.03 & 78.34  \\ 
& \textit{00873} & - & - & - & 96.87 & 80.24 \\ 
& \textit{00877} & - & - & - & 81.51 & 88.86 \\
& \textit{00890} & - & - & - & 95.25 & 62.82 \\ 
\cmidrule{2-7}
& Overall & - & - & - & \textbf{86.18} & 77.55 \\
 \midrule
\multirow{8}{*}{\rotatebox[origin=c]{90}{HOV-SG (ours)}} & \textit{00824} & 1.0 & 1 & 1 & 81.20 & 80.00 \\ 
& \textit{00829} & 1.0 & 1 & 1 & 88.81 & 88.02 \\ 
& \textit{00843} & 1.0 & 2 & 2 & 88.54 &87.10 \\ 
& \textit{00861} & 1.0 & 2 & 2 & 76.41 & 89.95 \\ 
& \textit{00862} & 1.0 & 3 & 3 & 72.65 & 76.10  \\ 
& \textit{00873} & 1.0 & 2 & 2 & 95.63 & 67.71 \\ 
& \textit{00877} & 1.0 & 2 & 2 & 74.82 & 92.30 \\ 
& \textit{00890} & 1.0 & 2 & 2 & 94.75 & 87.55 \\ 
\cmidrule{2-7}
& Overall & 1.0 & - & - & 84.10 & \textbf{83.59} \\

\bottomrule
\end{tabular}
\footnotesize
Evaluation of the floor and room segmentation: We provide the number of correctly predicted floors using a threshold of 0.5m. Room segmentation precision (P) and recall (R) are calculated based on the metric provided by Hydra~\cite{hughes2022hydra}.
\end{threeparttable}
\label{tab:floor-region-suppl}
\color{black}
\end{table}

\subsection{Room Classification}
\label{sec:room-classification-suppl}

\begin{table}[h]
\color{black}
\centering
\scriptsize
\caption{\textsc{Semantic Room Classification Results (HM3DSem)}}
\setlength\tabcolsep{6.7pt}
\begin{threeparttable}
\begin{tabular}{ll|ccc}
 \toprule
\multicolumn{2}{l}{Room Identification Method}  &  Scene & $\text{Acc}_{=}$\,[\%] & $\text{Acc}_{\approx}$\,[\%] \\
\midrule
\multirow{20}{*}{\rotatebox[origin=c]{90}{Privileged Methods}} & & \textit{00824} & 70.00 & 90.00 \\
& & \textit{00829} & 71.43 & 100.0 \\
& & \textit{00842} & 61.54 & 69.23 \\
& GPT-3.5 w/ ground-truth & \textit{00861} & 58.33 & 70.83 \\
& object categories & \textit{00862} & 50.00 & 72.22 \\
& (privileged) & \textit{00873} & 81.82 & 90.91 \\
& & \textit{00877} & 69.23 & 76.92 \\
& & \textit{00877} & 72.73 & 81.82 \\
 \cmidrule{3-5}
& & \textit{Overall} & 66.89 & 81.49 \\
 \cmidrule{2-5}
& & \textit{00824} & 80.00 & 90.00  \\
& & \textit{00829} & 85.71 & 85.71\\
& & \textit{00843} & 76.93 & 84.61 \\
& GPT-4 w/ ground-truth & \textit{00861} & 75.00 & 79.17 \\
& object categories & \textit{00862} & 63.89 & 66.67 \\
& (privileged) & \textit{00873} & 90.91 & 90.91 \\ 
& & \textit{00877} & 84.62 & 84.62 \\
& & \textit{00890} & 81.81 & 92.31 \\
 \cmidrule{3-5}
& & \textit{Overall} & 79.86 & 84.25 \\
\midrule
\multirow{30}{*}{\rotatebox[origin=c]{90}{Unprivileged Methods}} & & \textit{00824} & 30.00 & 40.00 \\
& & \textit{00829} & 42.86 & 57.14 \\
& & \textit{00843} & 38.46 & 38.46 \\
& GPT-3.5 w/ predicted & \textit{00861} & 16.67 & 25.00 \\
& object categories & \textit{00862} & 19.44 & 25.00 \\
& (unprivileged) & \textit{00873} & 45.45 & 63.64 \\ 
& & \textit{00877} & 07.69 & 30.77 \\
& & \textit{00877} & 27.27 & 63.64 \\
 \cmidrule{3-5}
& & \textit{Overall} & 28.48 & 42.95 \\
\cmidrule{2-5}
& & \textit{00824} & 70.00 & 80.00 \\
& & \textit{00829} & 71.43 & 71.43 \\
& & \textit{00843} & 53.85 & 53.85 \\
& GPT-4 w/ predicted & \textit{00861} & 45.83 & 45.83 \\
& object categories & \textit{00862} & 44.44 & 50.00 \\
& (unprivileged) & \textit{00873} & 72.73 & 81.81 \\ 
& & \textit{00877} & 53.85 & 53.85 \\
& & \textit{00890} & 63.63 & 63.63\\
 \cmidrule{3-5}
& & \textit{Overall} & \underline{59.47} & \underline{62.55} \\
\cmidrule{2-5}
& \multirow{9}{*}{View embeddings (ours)} & \textit{00824} & 80.00 & 90.00 \\
& & \textit{00829} & 85.71 & 100.0 \\
& & \textit{00842} & 69.23 & 76.92 \\
& & \textit{00861} & 54.17 & 79.17 \\
& & \textit{00862} & 63.89 & 83.33 \\
& & \textit{00873} & 90.91 & 90.91 \\
& & \textit{00877} & 61.54 & 61.54 \\
& & \textit{00877} & 81.82 & 90.91 \\
 \cmidrule{3-5}
& & \textit{Overall} & \textbf{73.93} & \textbf{84.10} \\
\bottomrule
\end{tabular}
\footnotesize
The table shows the room classification performance of our method (view embeddings) and two baselines (at the top) on HM3DSem. The baselines utilize GPT-3.5 / GPT-4 for labeling the rooms based on either ground-truth objects (masks) and categories or on predicted masks and categories. We consider two different evaluation criteria: $\text{Acc}_{=}$ measures whether the exact text-wise room category was predicted while $\text{Acc}_{\approx}$ measures correct room labels based on qualitative human evaluation.
\end{threeparttable}
\label{tab:habitat-room-eval-suppl}
\end{table}

\color{black}
As presented in Table~\ref{tab:habitat-room-eval-suppl}, we provide additional scene-wise results across the set of eight scenes in HM3DSem~\cite{habitat23semantics} that complement the results given in Table~\ref{tab:habitat-room-eval}. We note that there is only a single scene in which the privileged GPT-3.5 baseline outperforms our proposed view embedding method (\textit{00877}). The na\"ive baseline operating on predicted object categories is significantly outperformed across all scenes, which is mostly due to under-segmentation and wrongly predicted top-1 object categories.
\color{black}

\noindent\rebuttal{\textbf{Prompts:} For the GPT-based methods, we used the following prompts to derive room types based on the contained object names. The content for ``system'' is in {\color{prompt-gray}gray}. The content for ``user'' is in \command{green}. And the content for ``assistant'' is \colorbox{highlight}{highlighted}. During test time, the user needs to provide \{objects\} and \{room\_types\} which are the object names in the room and the pre-defined set of room types to select from. The prompt is shown below:}
\lmp{
    \scriptsize
    \prompt{system: You are a room type detector. You can infer a room type based on a list of objects.}\\
    \command{Q: The list of objects contained in this room are: bed, wardrobe, chair, sofa. What is the room type? Please just answer the room name.}\\
    \colorbox{highlight}{A: bedroom}\\
    \command{Q: The list of objects contained in this room are: tv, table, chair, sofa. Please pick the most matching room type from the following list: living room, bedroom, bathroom, kitchen. What is the room type? Please just answer the room name.}\\
    \colorbox{highlight}{A: living room}\\
    \command{Q: The list of objects contained in this room are: \{objects\}. Please pick the most matching room type from the following list: \{room\_types\}. What is the room type? Please just answer the room name.}\\
}

\noindent\rebuttal{\textbf{Room Categories:} We manually label the GT room regions with the following 14 room types: \texttt{living room, dining room, kitchen, bathroom, bedroom, dressing room, combined kitchen and living room, entryway, basement, laundry room, office, empty room, hallway, closet}.}

\begin{table}[h]
\centering
\scriptsize
\caption{\textsc{Object Retrieval from Language Queries (HM3DSem)}}
\setlength\tabcolsep{3.7pt}
\begin{threeparttable}
\begin{tabular}{ll|ccccc}
 \toprule
Query Type &Method &  Scene & \#\,Floors & \#\,Regions & \#\,Trials & SR$_{10}$[\%] \\
\midrule
\multirow{20}{*}{(\texttt{o}, \texttt{r}, \texttt{f})} & \multirow{9}{*}{ConceptGraphs} & \textit{00824} & 1 & 10 & 33 & 33.33 \\
& & \textit{00829} & 1 & 7 & 20 & \textbf{65.00} \\
& & \textit{00843} & 2 & 13 & 26 & 03.85 \\
& & \textit{00861} & 2 & 24 & 55 & 01.82  \\
& & \textit{00862} & 3 & 36 & 90 & \textbf{21.11} \\
& & \textit{00873} & 2 & 11 & 28 & 10.71 \\
& & \textit{00877} & 2 & 13 & 32 & 09.38\\
& & \textit{00890} & 2 & 11 & 41 & 04.88\\
 \cmidrule{3-7}
  & & \textit{Overall} & - & - & 40.63 & 16.31 \\
  \cmidrule{2-7}
 & & \textit{00824} & 1 & 10 & 33 & \textbf{57.57} \\ 
& & \textit{00829} & 1 & 7 & 20 & 45.00 \\ 
& & \textit{00843} & 2 & 13 & 26 & \textbf{34.62} \\ 
& & \textit{00861} & 2 & 24 & 55 & \textbf{25.45} \\ 
& {HOV-SG (ours)}  & \textit{00862} & 3 & 36 & 90 & \textbf{21.11} \\ 
& w/ OVSeg & \textit{00873} & 2 & 11 & 28 & \textbf{14.29} \\ 
& & \textit{00877} & 2 & 13 & 32 & \textbf{25.00} \\ 
& & \textit{00890} & 2 & 11 & 41 & \textbf{21.95} \\ 
 \cmidrule{3-7}
 & & \textit{Overall} &  &  & 40.63 & \textbf{28.00} \\
 \midrule\midrule
\multirow{20}{*}{(\texttt{o}, \texttt{r})} & \multirow{9}{*}{ConceptGraphs} & \textit{00824} & 1 & 10 & 33 & 33.33 \\
& & \textit{00829} & 1 & 7 & 20 & \textbf{65.00} \\
& & \textit{00843} & 2 & 13 & 23 & 34.78 \\
& & \textit{00861} & 2 & 24 & 46 & 19.57 \\
& & \textit{00862} & 3 & 36 & 67 & \textbf{26.98} \\
& & \textit{00873} & 2 & 11 & 25 & \textbf{30.00} \\
& & \textit{00877} & 2 & 13 & 24 & 25.00 \\
& & \textit{00890} & 2 & 11 & 41 & \textbf{\underline{21.95}} \\
 \cmidrule{3-7}
  & & \textit{Overall} & - & - & 34.88 & 29.26 \\
  \cmidrule{2-7}
 & & \textit{00824} & 1 & 10 & 33 & \textbf{57.58} \\
& & \textit{00829} & 1 & 7 & 20 & 45.00 \\
& & \textit{00843} & 2 & 13 & 23 & \textbf{39.13} \\
& & \textit{00861} & 2 & 24 & 46 & \textbf{30.43} \\
& {HOV-SG (ours)}   & \textit{00862} & 3 & 36 & 67 & 20.63 \\
& & \textit{00873} & 2 & 11 & 25 & 20.00 \\ 
& & \textit{00877} & 2 & 13 & 24 & \textbf{33.33} \\
& & \textit{00890} & 2 & 11 & 41 & \textbf{\underline{21.95}} \\
 \cmidrule{3-7}
 & & \textit{Overall} & - & - & 34.88 & \textbf{31.48} \\
\bottomrule
\end{tabular}
\footnotesize
Evaluation over 20 frequent distinct object categories in terms of the top-5 accuracy. A match is counted as a success when the $\text{IoU}>0.1$ between predicted object and ground truth. The floor and room counts refer to the ground-truth labels. The number of trials is lower for \texttt{(o,r)} compared to \texttt{(o,r,f)} because we observe a higher number of query duplicates whenever we drop the floor specification. The 20 categories evaluated are: \textit{picture, pillow, door, lamp, cabinet, book, chair, table, towel, plant, sink, stairs, bed, toilet, tv, desk, couch, flowerpot, nightstand, faucet}.
\end{threeparttable}
\label{tab:habitat-retrieval-cg}
\end{table}

\subsection{Language-Grounded Navigation with Long Queries}

In order to support our proposed hierarchical segregation of the environment, we present another comparison with ConceptGraphs~\cite{conceptgraphs}.
To do so, we compare the object retrieval from language queries performance to demonstrate the efficacy of hierarchically decomposing scenes. We draw this comparison by augmenting ConceptGraphs to also work with room and floor queries.
For both HOV-SG as well as ConceptGraphs, we decompose the original query via GPT3.5 parsing as before. Using this, we obtain text variables stating the requested floor name, room name, and object name. Since the floor segmentation of HOV-SG consistently showed 100\% accuracy, we directly provide ConceptGraphs with that information. Our augmentation of ConceptGraphs allows us to implicitly identify potential target rooms and objects: We compute the cosine similarity between the set of all object embeddings and the queried room text. Similarly, we compute the cosine similarity between the set of all objects and the queried object name. We combine these two similarities by taking the product of those scores per object to identify the most probable objects. This allows ConceptGraphs to draw answers at the floor level and room level. The remaining details of this evaluation are detailed in the main manuscript. 

The results in Table~\ref{tab:habitat-retrieval-cg} regarding object-room-floor queries demonstrate a significant performance improvement of 11.69\% when using HOV-SG compared to ConceptGraphs. We observe that ConceptGraphs struggles with larger scenes and under-segmentation of the produced maps, which often makes finding the object in question hard. Regarding the object-room queries, the drawbacks of ConceptGraphs are not as apparent because the search domain is significantly larger. Still, HOV-SG shows a 2.2\% advantage over ConceptGraphs. In general, erroneous room segmentations produced by HOV-SG make finding the object in question hard, which remains subject to future work.

\subsection{Graph Representation on HM3DSem}
In the following, we also show the produced hierarchical 3D scene graphs on the set of 8 scenes we evaluated in Fig.~\ref{fig:hm3dsem-sg-vis}. Each distinct object is colored with a different color and the ground truth floor surface is underlayed for easier visibility. The blue nodes denote rooms and its links to the objects denote the object-room associations. The edges among the yellow nodes and the blue nodes show the association between rooms and floors. For clear visualization, we do not visualize the root node that connects multiple floors. We reject certain objects for visualization based on their top-1 predicted object category (out of 1624 categories). Any categories containing sub-strings of the following have not been
visualized: wall, floor, ceiling, paneling, banner, overhang. All other predicted object categories are shown. Remarkably, this procedure removed the fair majority of ceilings, walls, etc., which confirms the accuracy of the top-1 predicted open-vocabulary object labels. Nonetheless, future work should address the problem of over- and under-segmentation in these maps. Coping with multiple overlapping masks produced during iterative mask merging is still an open question. While having multiple overlapping masks per point drives the recall in semantic retrieval, this does not produce visually appealing maps. In general, one could argue that depending on the language query at hand different concepts are requested. In case of a query such as  \textit{"Find the sofa"}, one would like to obtain the mask that encloses the whole sofa. On the other hand, if the query comes in the form of \textit{"Find the cushion"} (on the sofa), we would want to singulate the cushion in question. This however is difficult when the sofa is masked as one, which would then be considered under-segmentation. Thus, we envision maps that can hold multiple overlapping object masks that could represent various sub-concepts. Essentially, this translates to an additional object hierarchy layer that decomposes objects into their parts.

\begin{figure}
\includegraphics[width=0.5\textwidth]{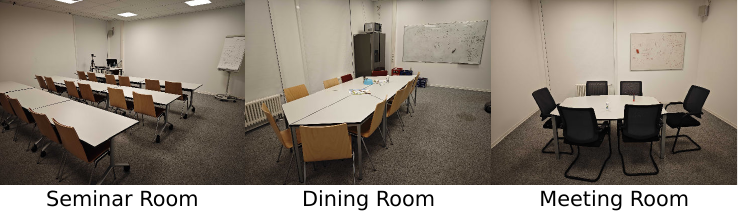}
\caption{\rebuttal{Camera images obtained in the ``seminar room'', ``dining room'', and ``meeting room''. We observe that these rooms all contain multiple chairs and large tables. In addition, the definition of the room types can be subjective.
This poses a considerable challenge when trying to differentiate among these rooms given their visual CLIP embeddings.}}
\label{fig:room_comparison}
\end{figure}

\begin{figure*}
\centering
\footnotesize
\setlength{\tabcolsep}{0.3cm}
{
\renewcommand{\arraystretch}{0.5}
\newcolumntype{M}[1]{>{\centering\arraybackslash}m{#1}}
\begin{tabular}{M{7.1cm}M{7.1cm}}
{\includegraphics[width=0.95\linewidth]{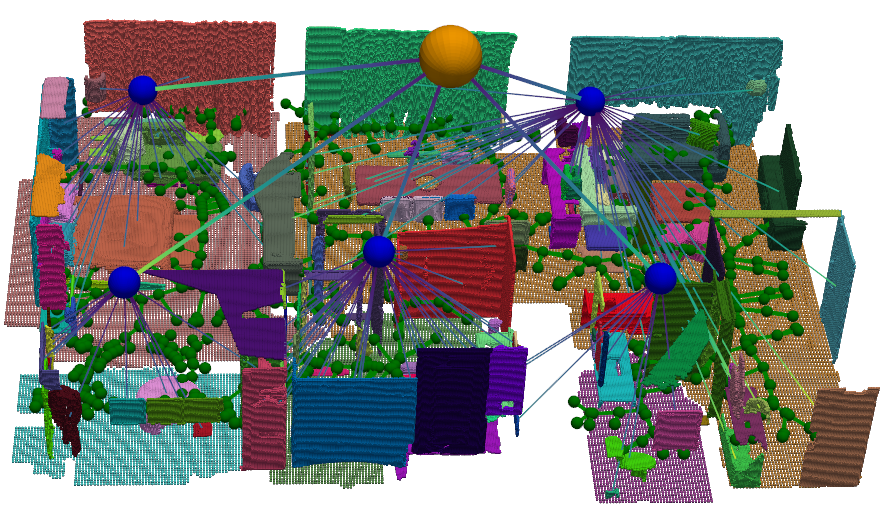}} & 
{\includegraphics[width=0.95\linewidth]{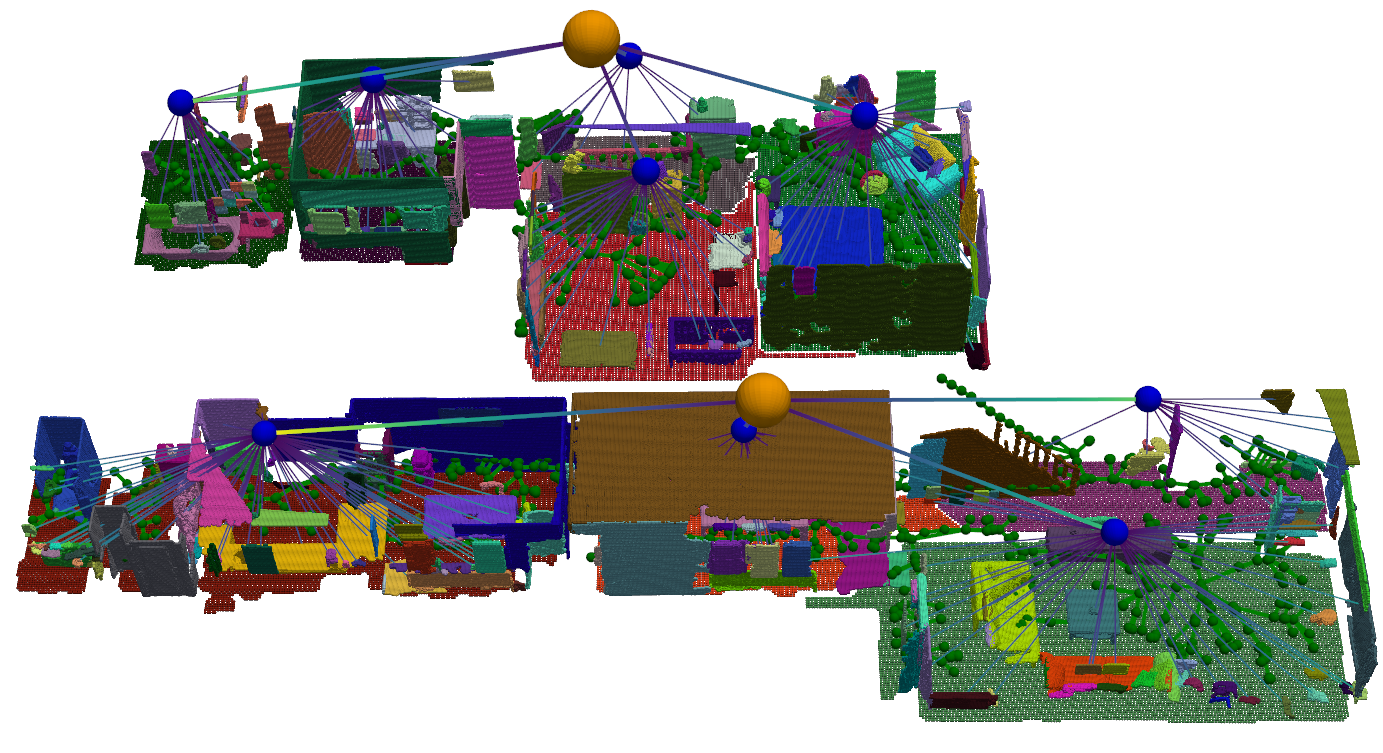}}
\\
\textit{Scene 00829 (1 floor)} & \textit{Scene 00890 (2 floors)} 
\\
\\
\\
\\
\\
{\includegraphics[width=0.95\linewidth]{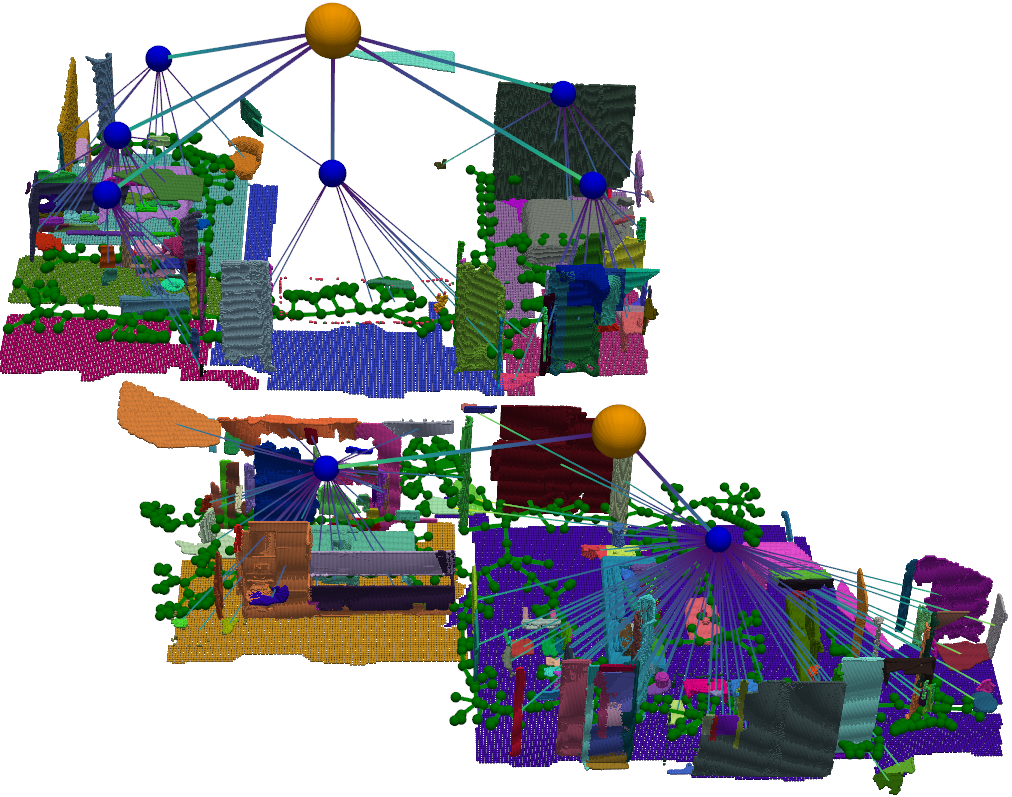}} & 
{\includegraphics[width=0.95\linewidth]{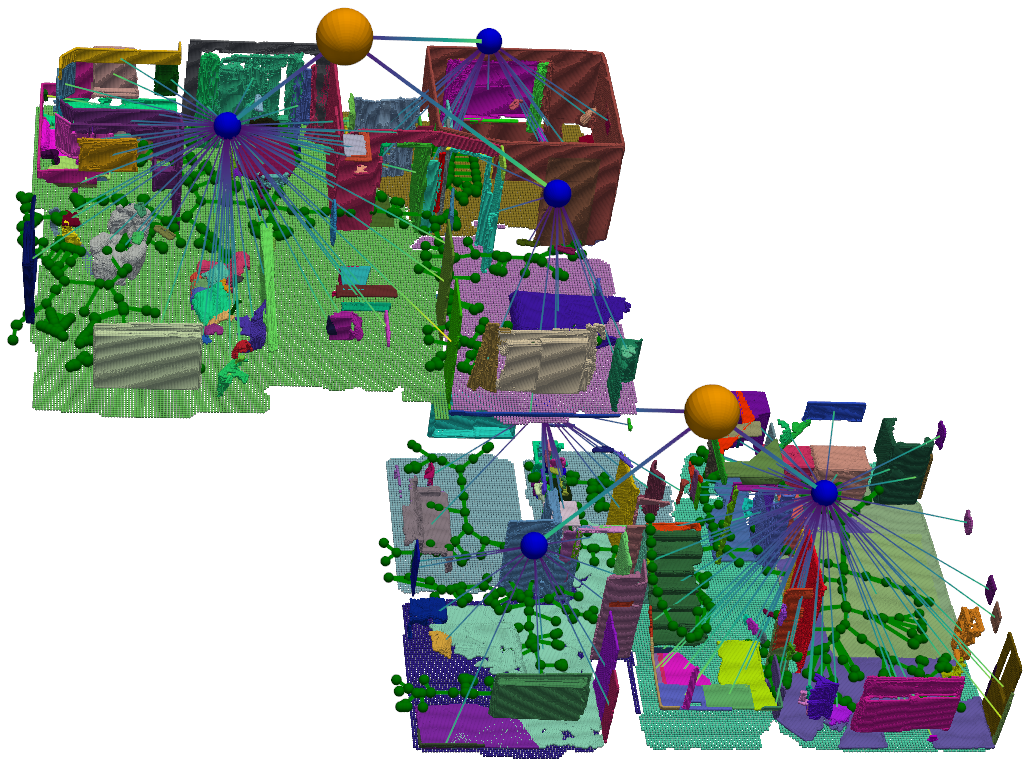}}
\\
\textit{Scene 00843 (2 floors)} & \textit{Scene 00877 (2 floors)}
\\
\\
\\
\\
\\
{\includegraphics[width=0.95\linewidth]{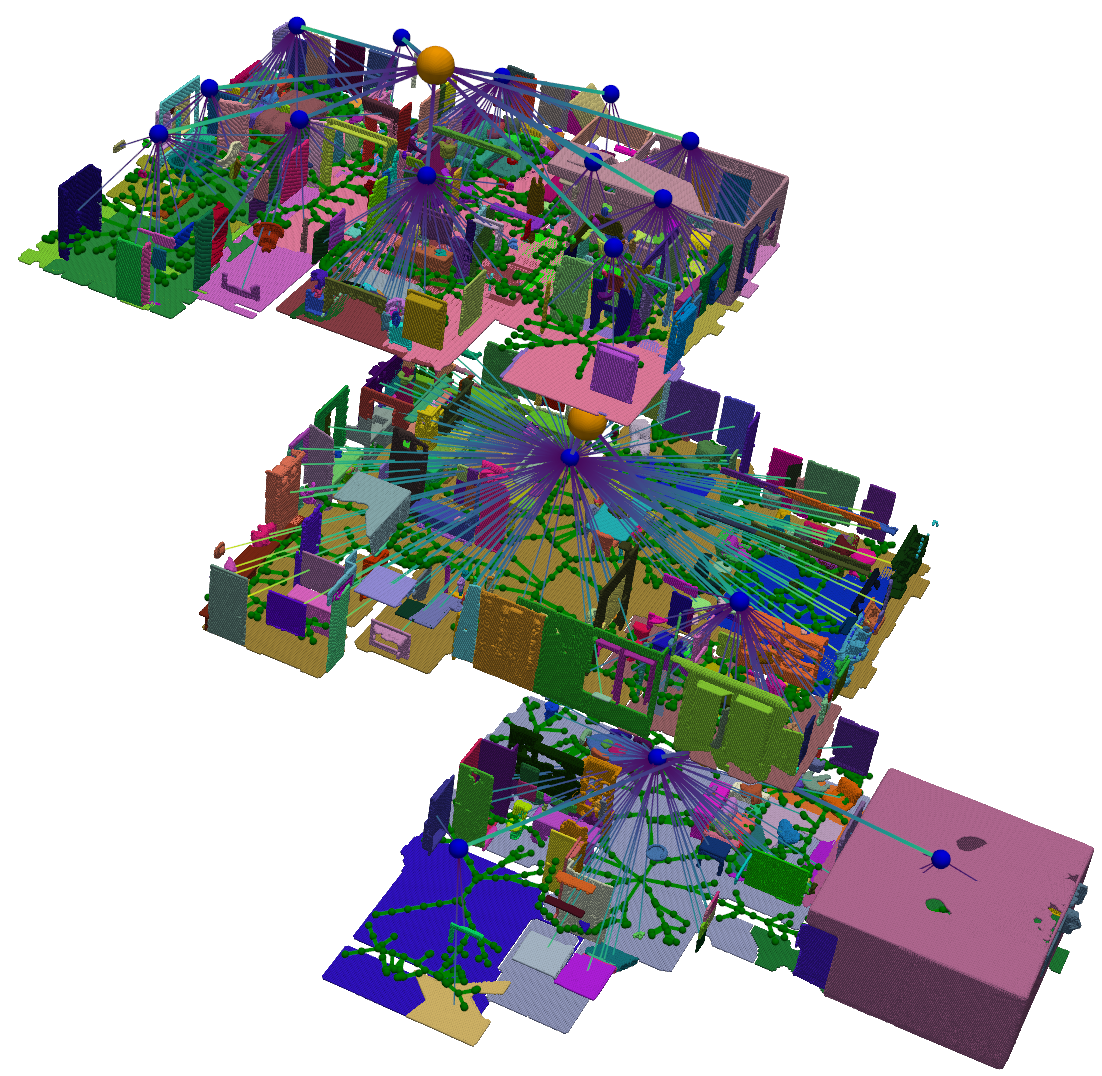}} & 
{\includegraphics[width=0.95\linewidth]{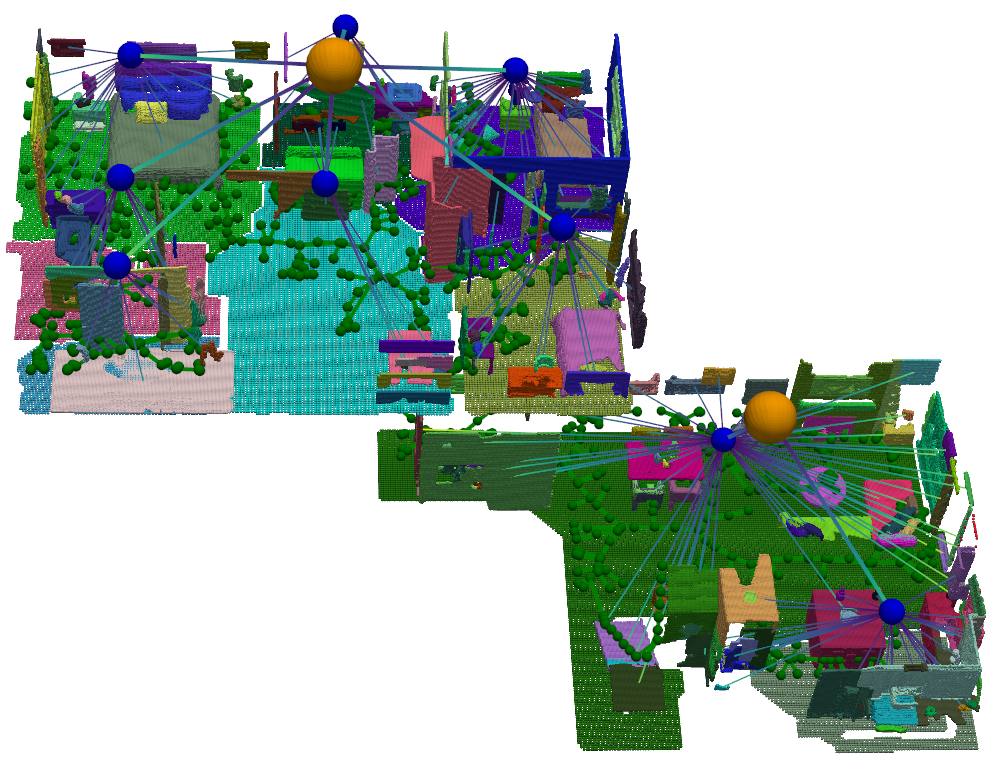}}
\\
\textit{Scene 00862 (3 floors)} & \textit{Scene 00873 (2 floors)}
\\
\\
\end{tabular}
}
\caption{We show a visualization of the hierarchical open-vocabulary scene graphs produced on HM3Dsem. To make the visualization more clear we do not show the root node connecting (multiple) floors. In addition, we underlay the ground-truth floor surface for easier visibility. We reject certain objects for visualization based on their top-1 predicted object category (out of 1624 categories). Any categories containing sub-strings of the following have not been visualized: \texttt{wall, floor, ceiling, paneling, banner, overhang}. All other predicted object categories are shown. Remarkably, this procedure removed the fair majority of ceilings, walls, etc., which confirms the accuracy of the top-1 predicted open-vocabulary object labels. Best viewed zoomed in.}
\label{fig:hm3dsem-sg-vis-suppl}

\end{figure*}

\subsection{Real-World Ambiguous Room Labels}
\rebuttal{As mentioned in Sec.~\ref{sec:exp_lang_nav_real_world}, one major failure reason is the challenge of differentiating room types with similar visual appearance and the subjectivity of the room definitions. Througout our real-world evaluation of \ours{}, we noticed that ``dining room'', ``seminar room'', and ``meeting room'' all contain many chairs and large tables as shown in Fig.~\ref{fig:room_comparison}.}

\subsection{Real-World Qualitative Navigation Results}
In Fig.~\ref{fig:real-world-trials}, we present three real-world trials that were executed with a Boston Dynamics Spot quadrupedal robot, which allowed us to traverse multi-floor environments safely. The trials are performed based on complex hierarchical language queries that specify the floor, the room, and the object to find. All hierarchical concepts relied on in these experiments are identified using our open-vocabulary HOV-SG pipeline. The top row in Fig.~\ref{fig:real-world-trials} shows the taken path (blue) from the start position (red) to the goal location (green). The following rows show the time-wise progression of the trial from top to bottom. The unique difficulty in these experiments is the typical office/lab environment with many similar rooms, which often produced similar room names. Having semantically varied rooms instead drastically simplifies these tasks. Nonetheless, as reported in the main manuscript, we reach real-world success rates of around 55\%. We also showed a subset of target objects in Fig.~\ref{fig:real_world_target_object}.

\begin{figure*}
\centering
\footnotesize
\setlength{\tabcolsep}{0.3cm}
{
\renewcommand{\arraystretch}{0.5}
\newcolumntype{M}[1]{>{\centering\arraybackslash}m{#1}}

\begin{tabular}{M{5.1cm}M{5.1cm}M{5.1cm}}
\textit{"Find the bean bag in the office on floor 1"} & \textit{"Find the coat in the office on floor 1"} & \textit{"Find the toilet in the bathroom on floor 2"}
\\
\\
{\includegraphics[width=\linewidth]{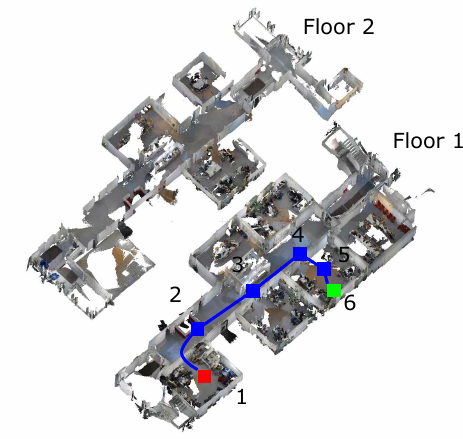}} & 
{\includegraphics[width=\linewidth]{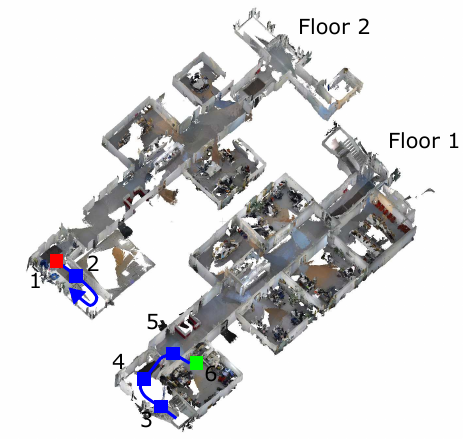}} & 
{\includegraphics[width=\linewidth]{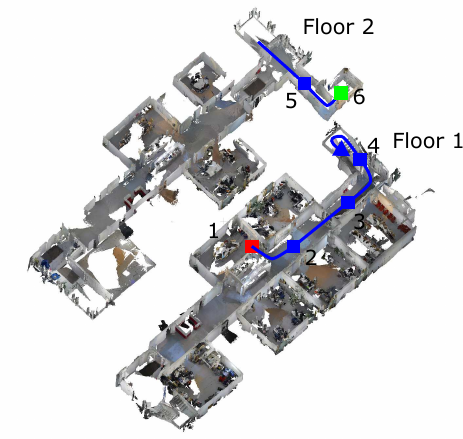}} 
\\
\\
{\includegraphics[width=\linewidth]{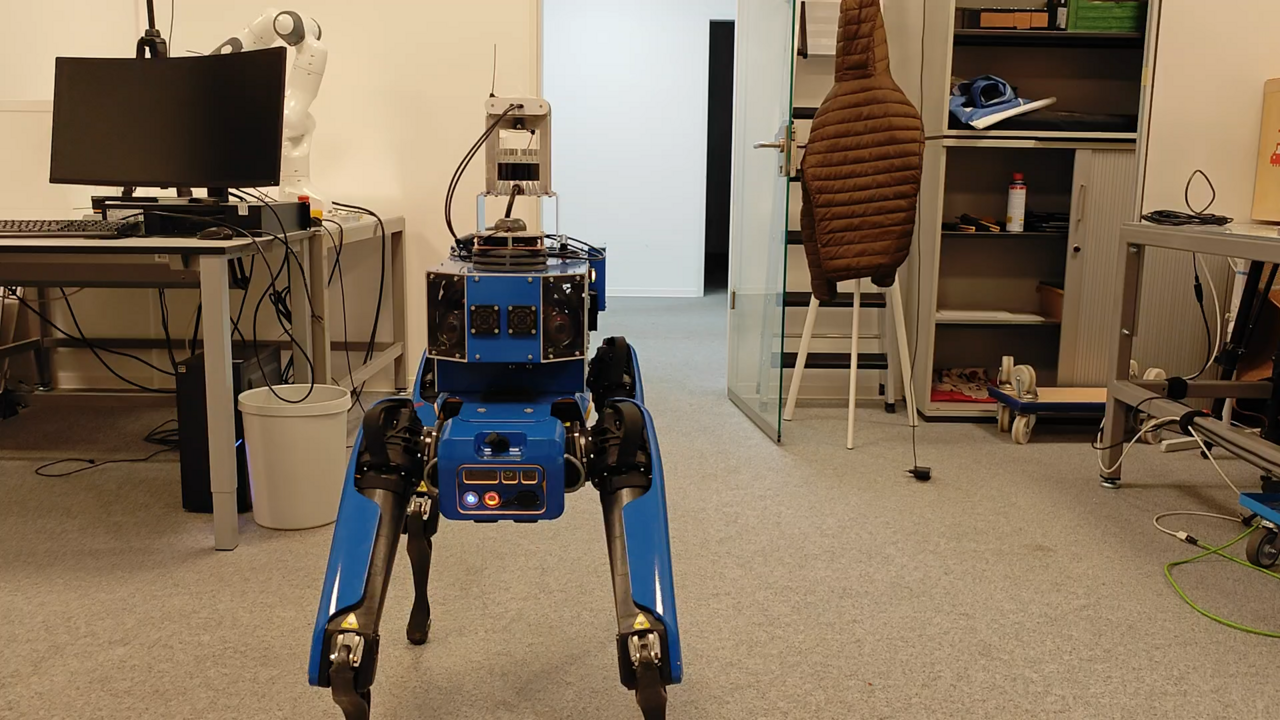}} & 
{\includegraphics[width=\linewidth]{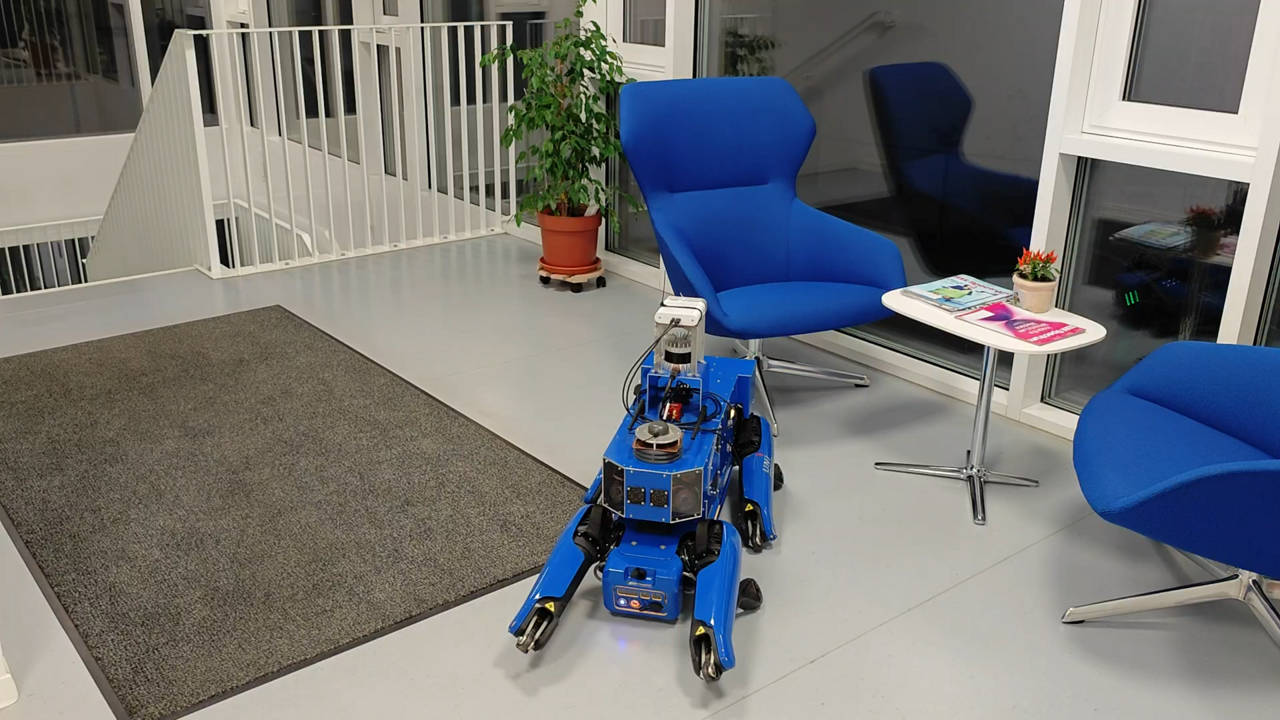}} & 
{\includegraphics[width=\linewidth]{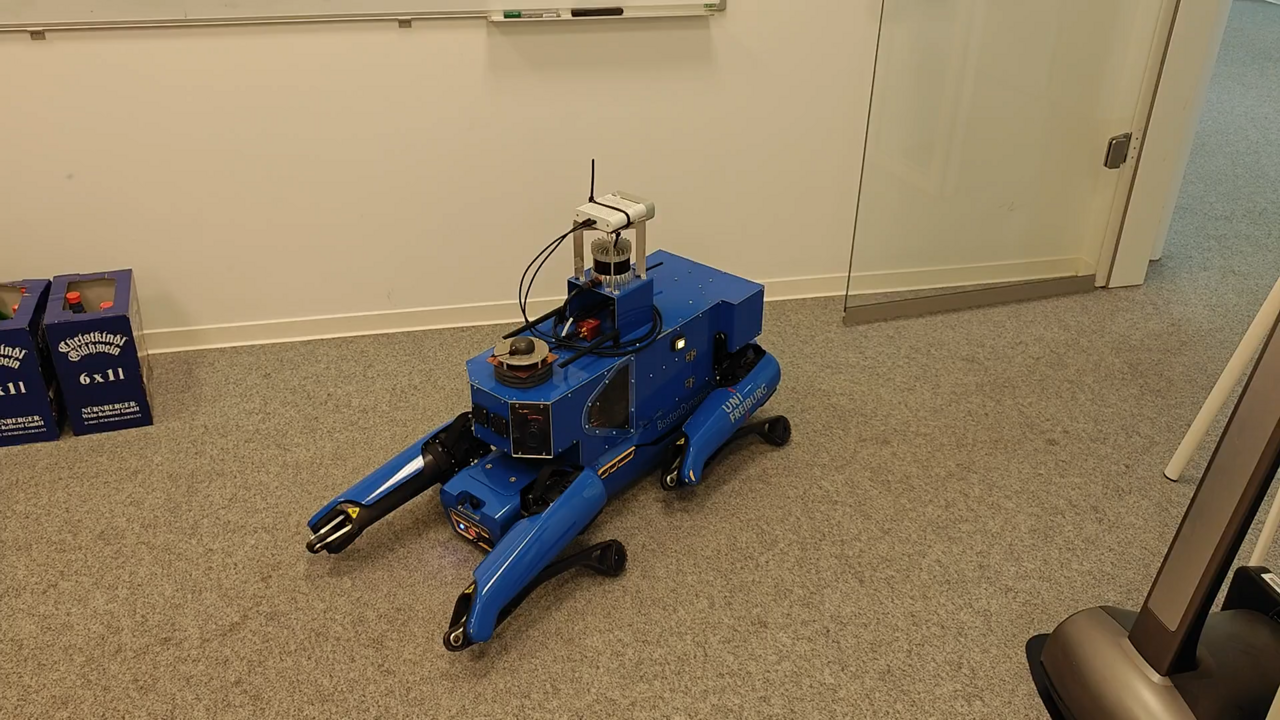}} 
\\
\\
{\includegraphics[width=\linewidth]{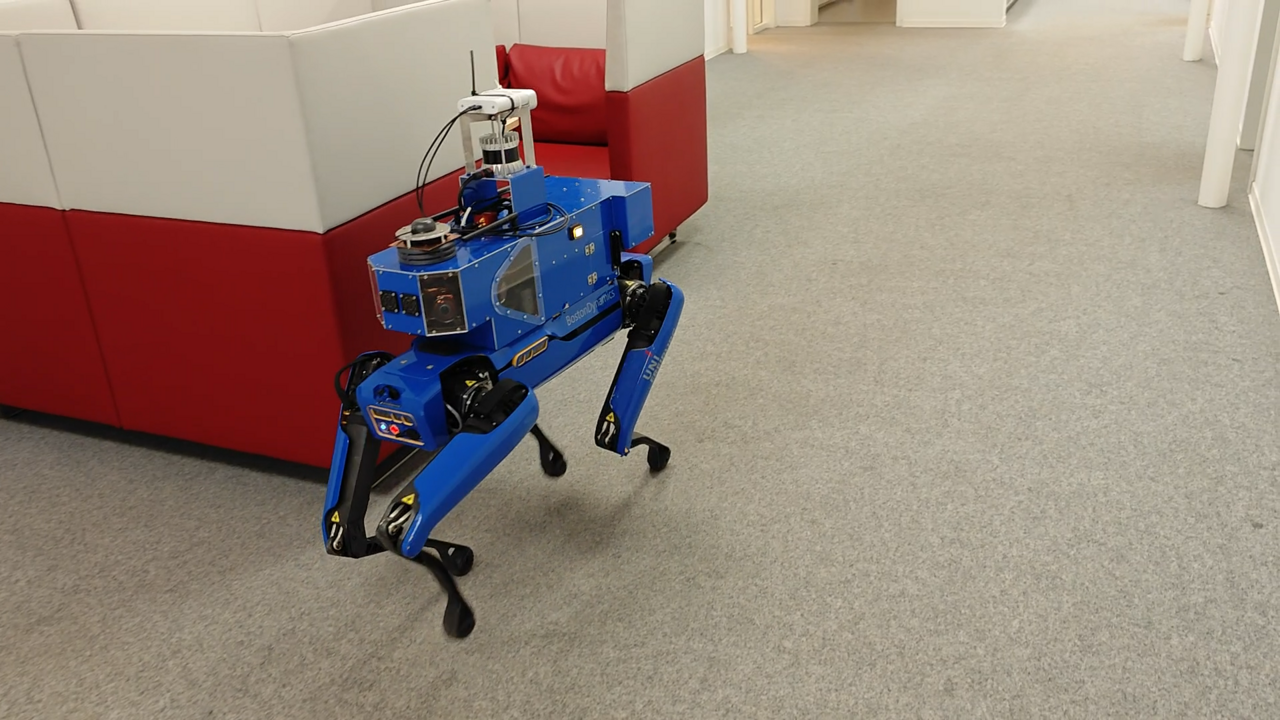}} & 
{\includegraphics[width=\linewidth]{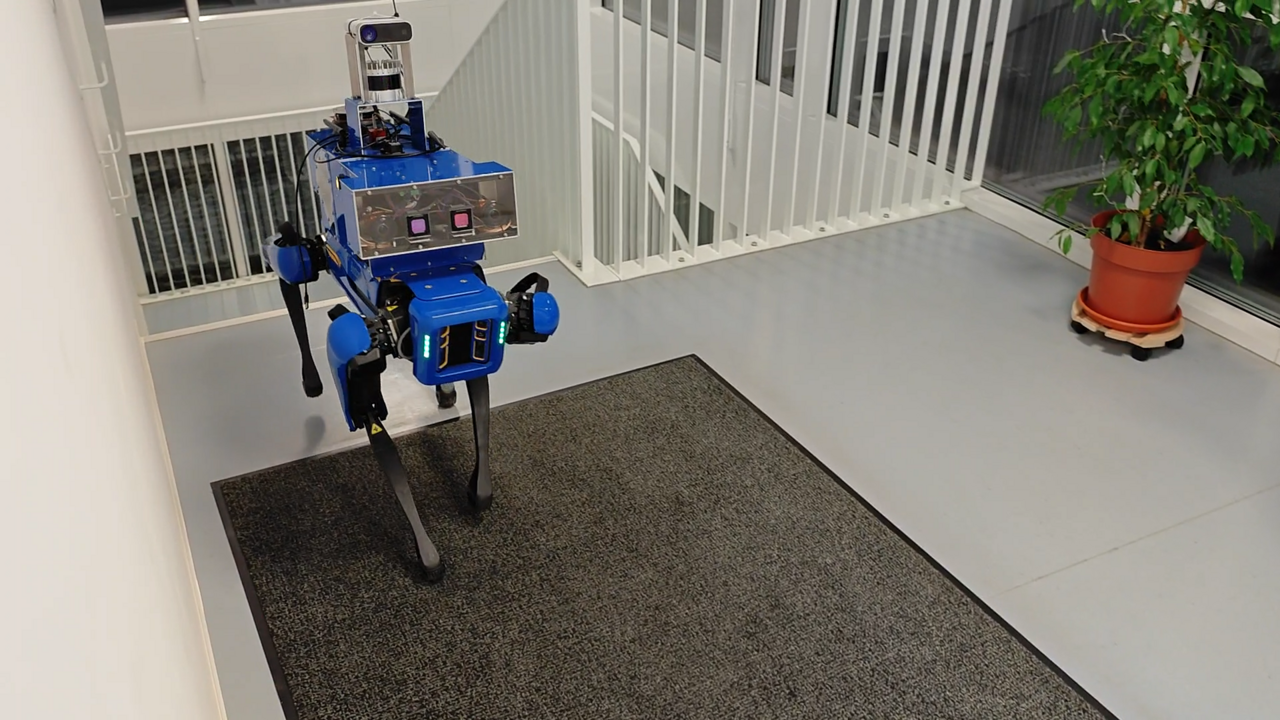}} & 
{\includegraphics[width=\linewidth]{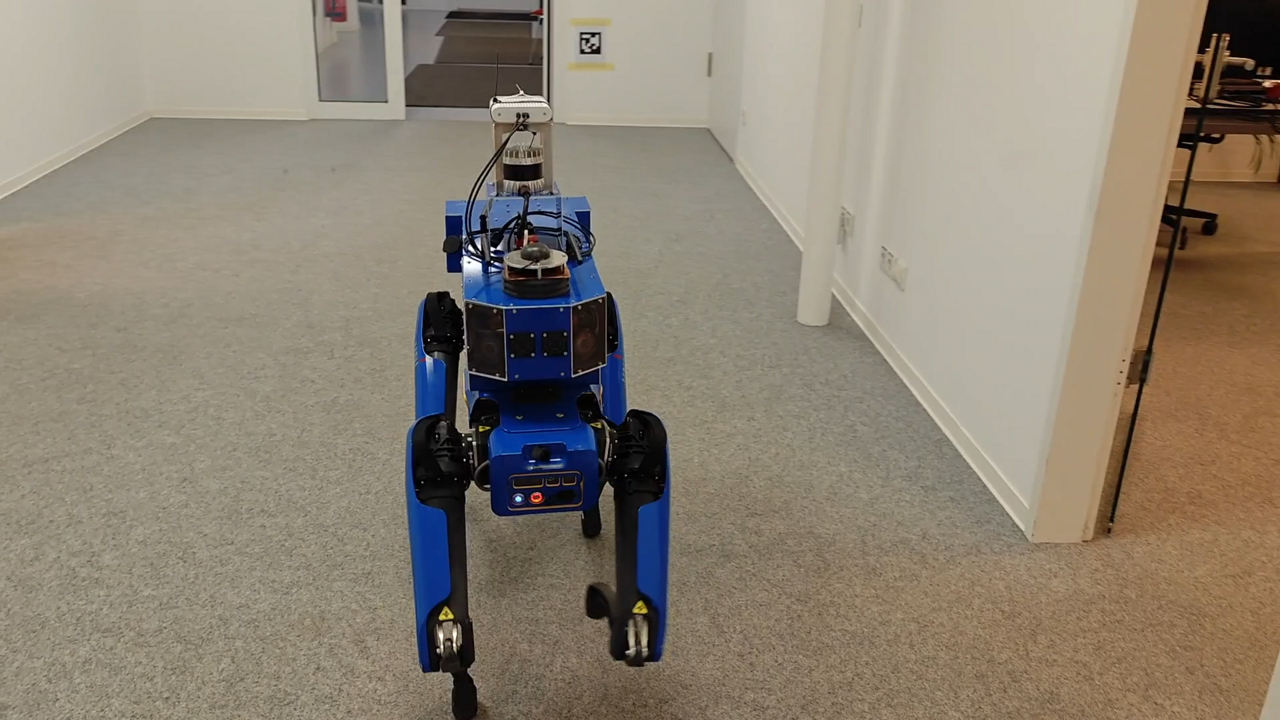}} 
\\
\\
{\includegraphics[width=\linewidth]{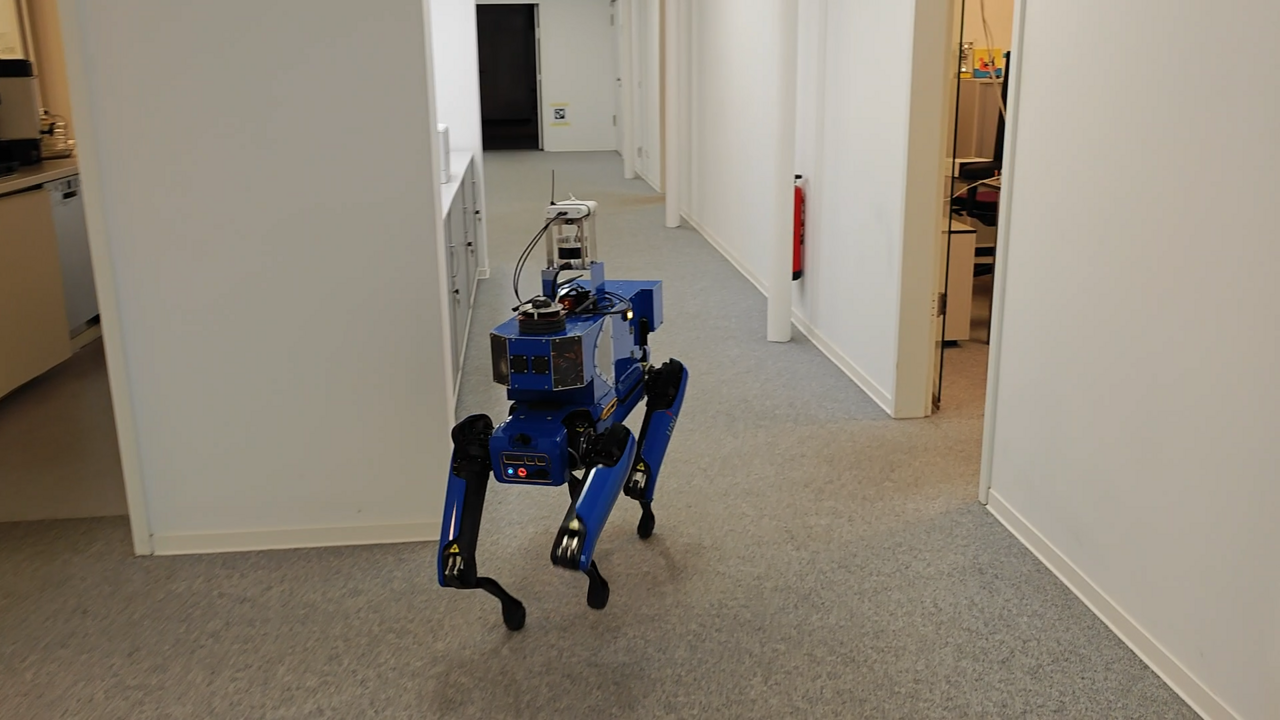}} & 
{\includegraphics[width=\linewidth]{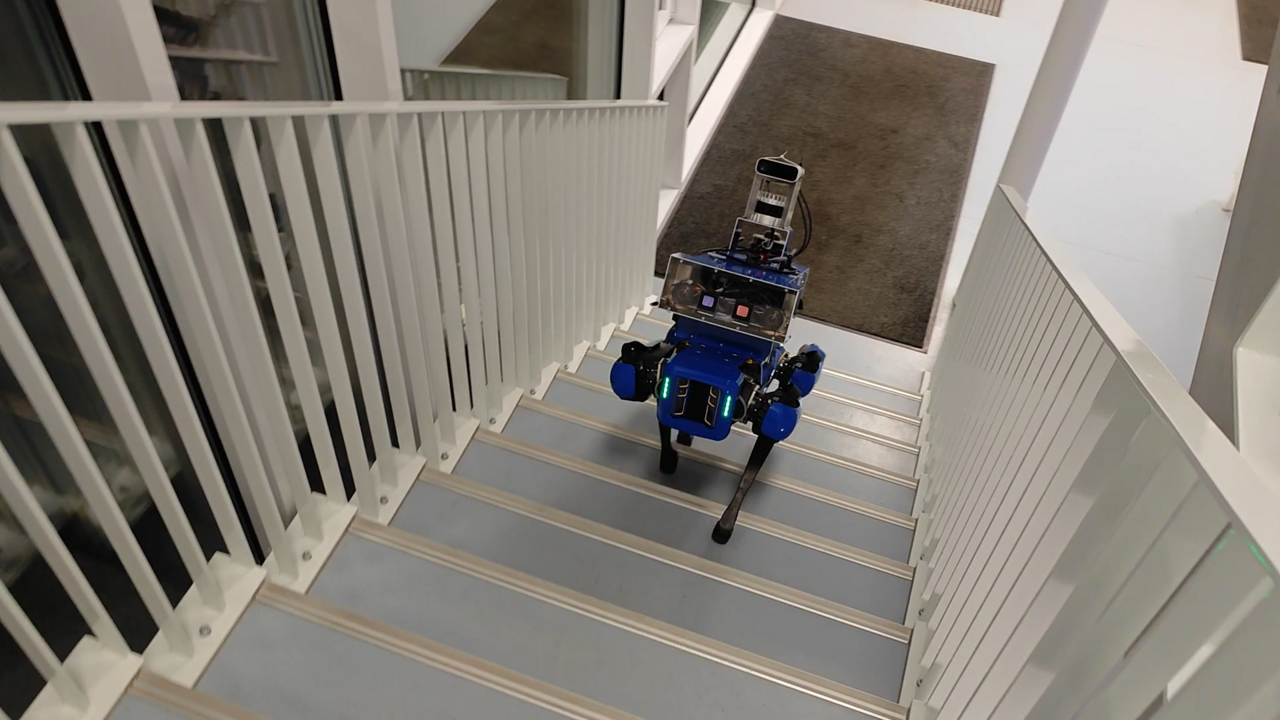}} & 
{\includegraphics[width=\linewidth]{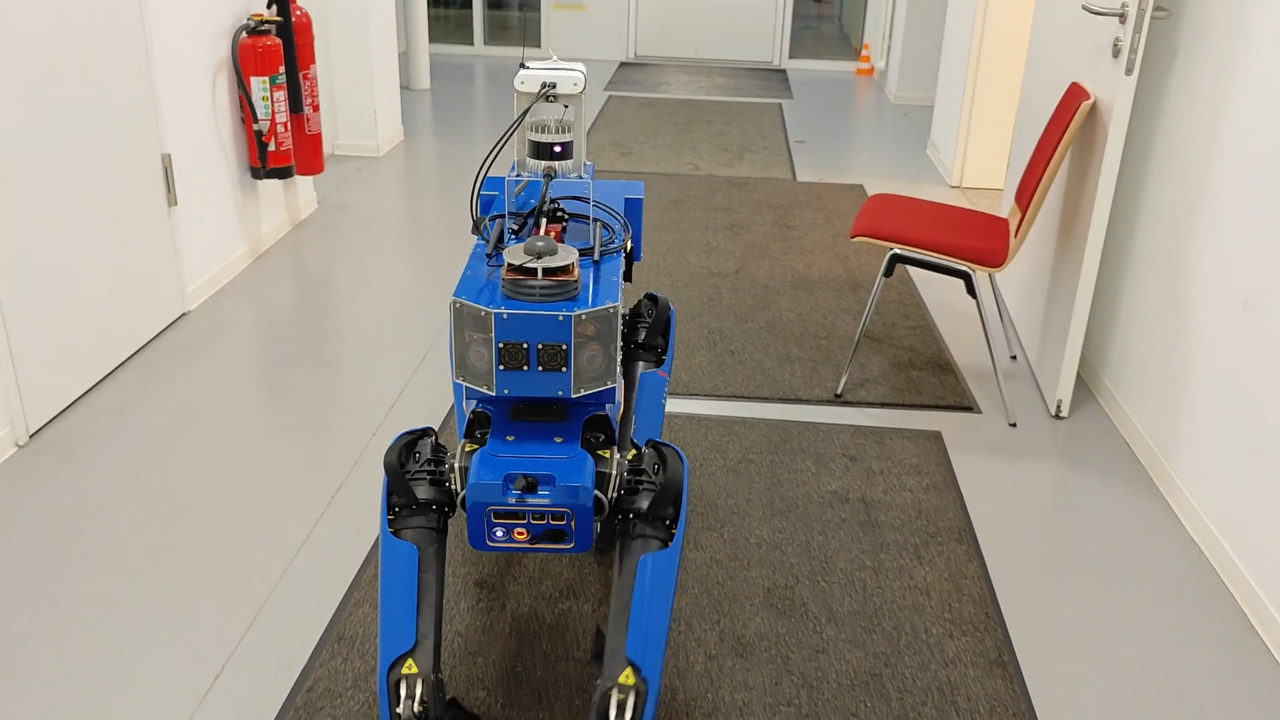}} 
\\
\\
{\includegraphics[width=\linewidth]{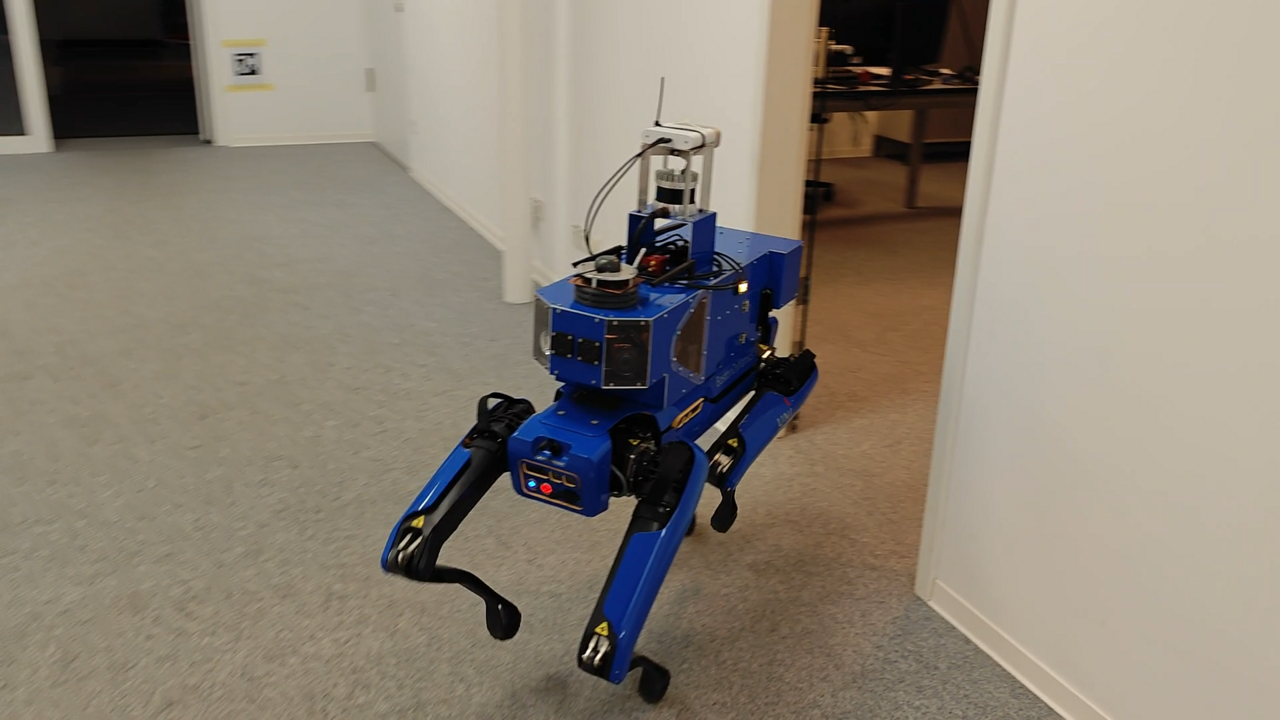}} & 
{\includegraphics[width=\linewidth]{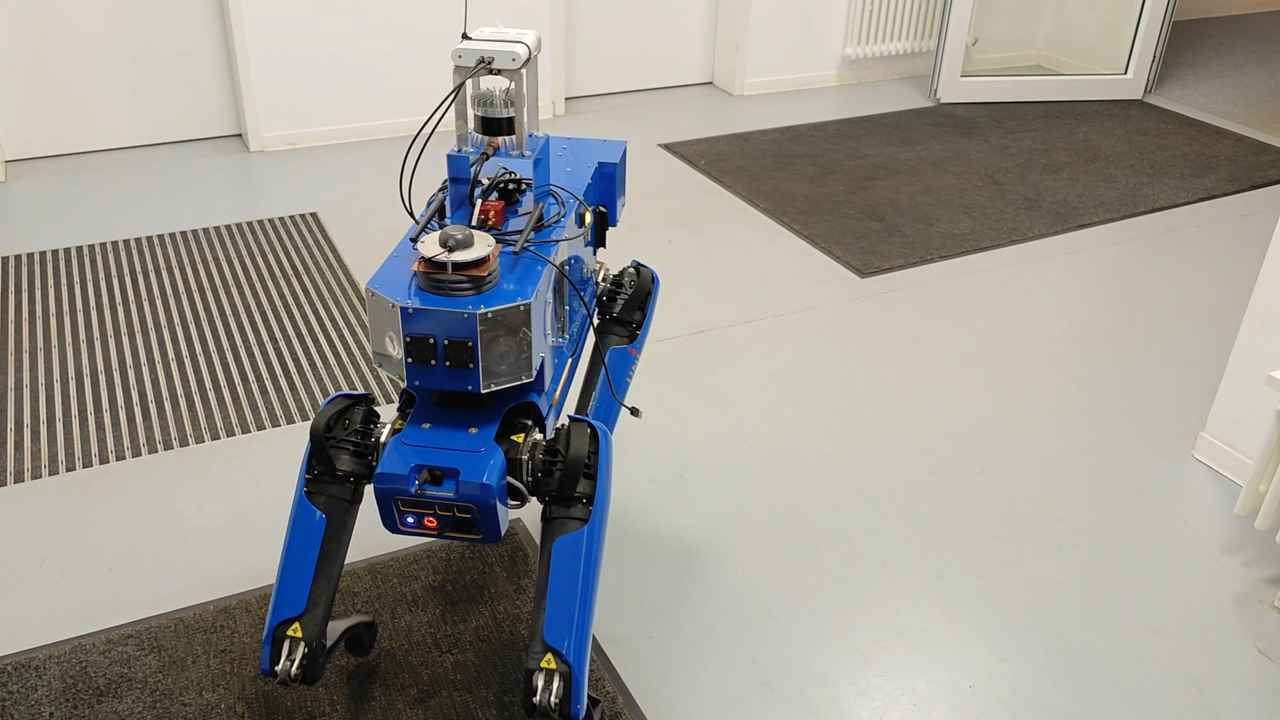}} & 
{\includegraphics[width=\linewidth]{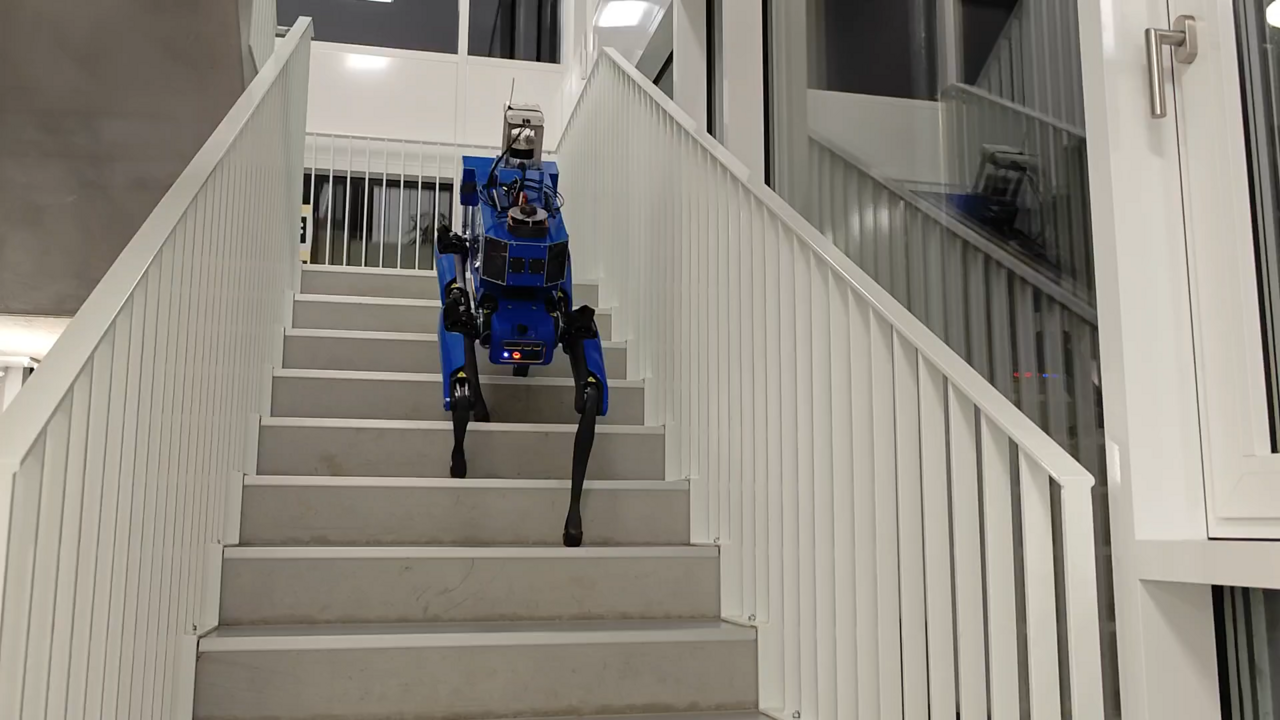}} 
\\
\\
{\includegraphics[width=\linewidth]{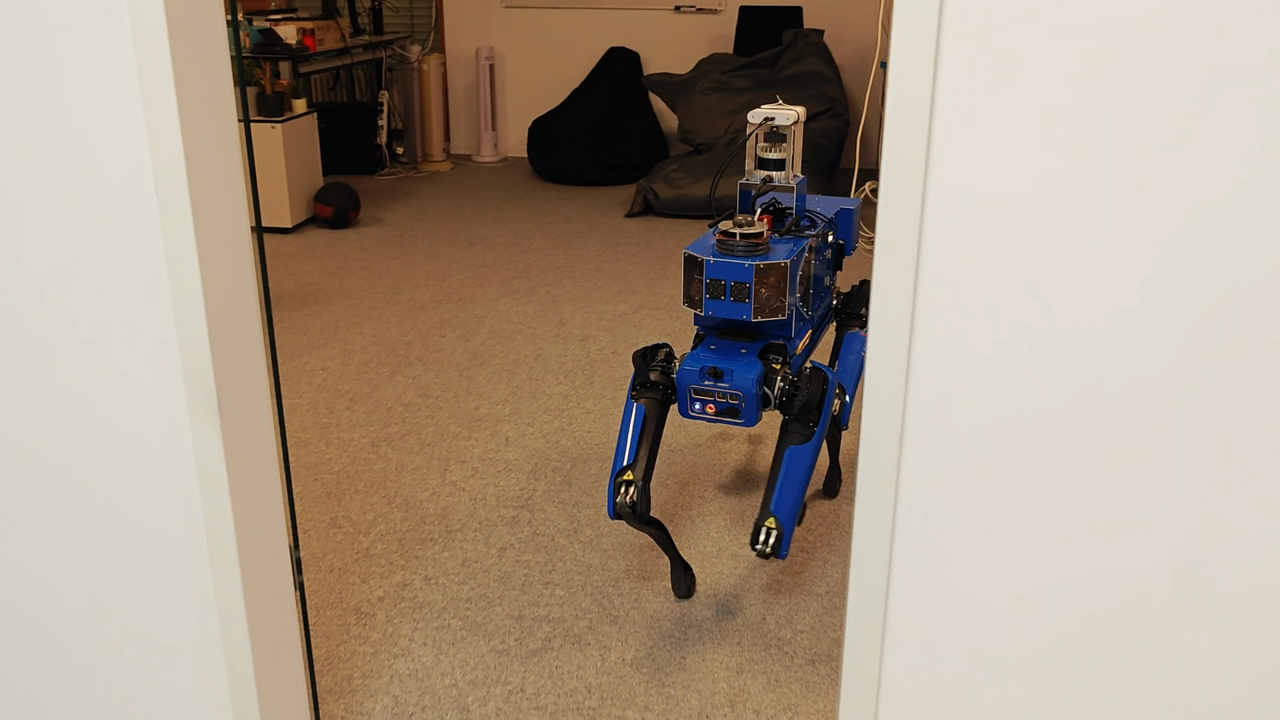}} & 
{\includegraphics[width=\linewidth]{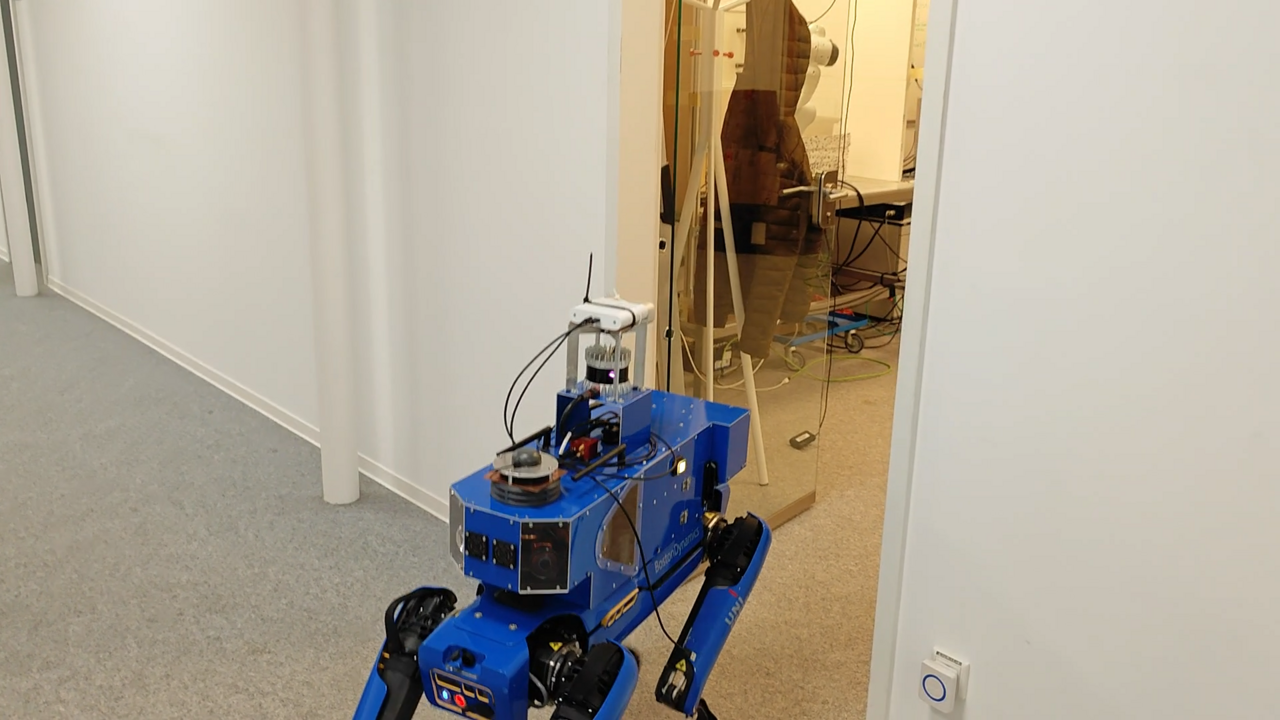}} & 
{\includegraphics[width=\linewidth]{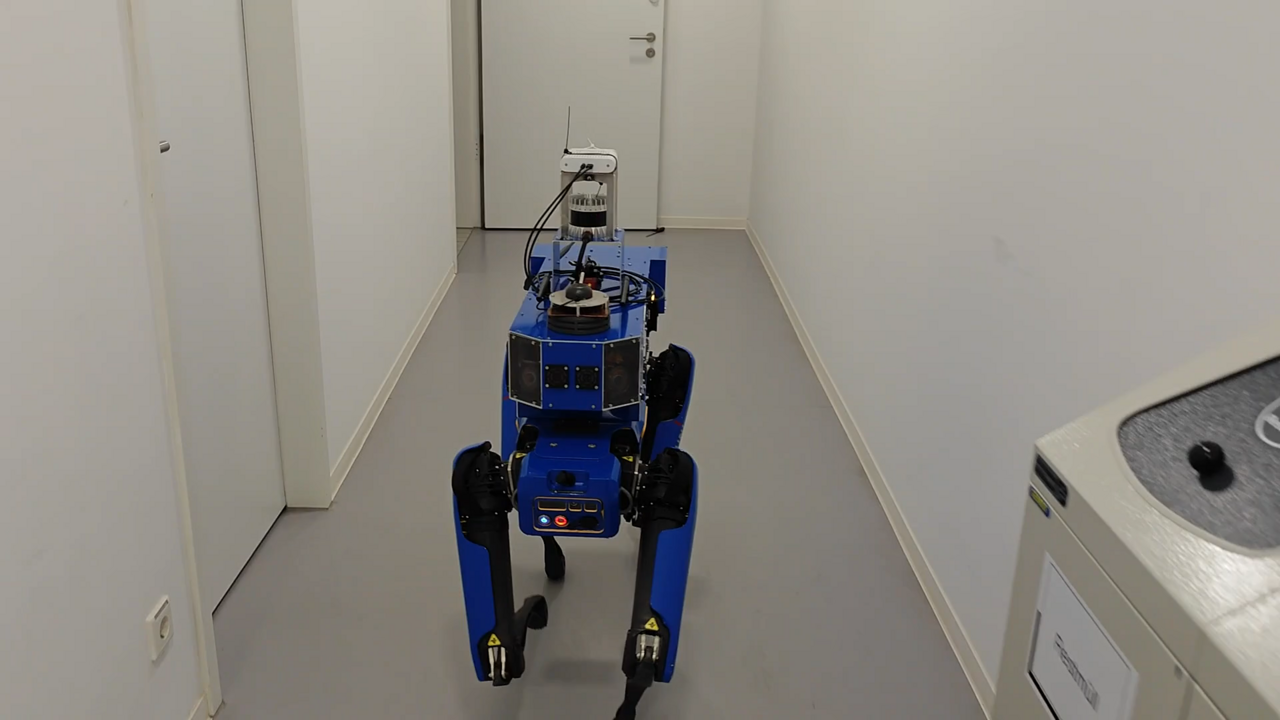}} 
\\
\\
{\includegraphics[width=\linewidth]{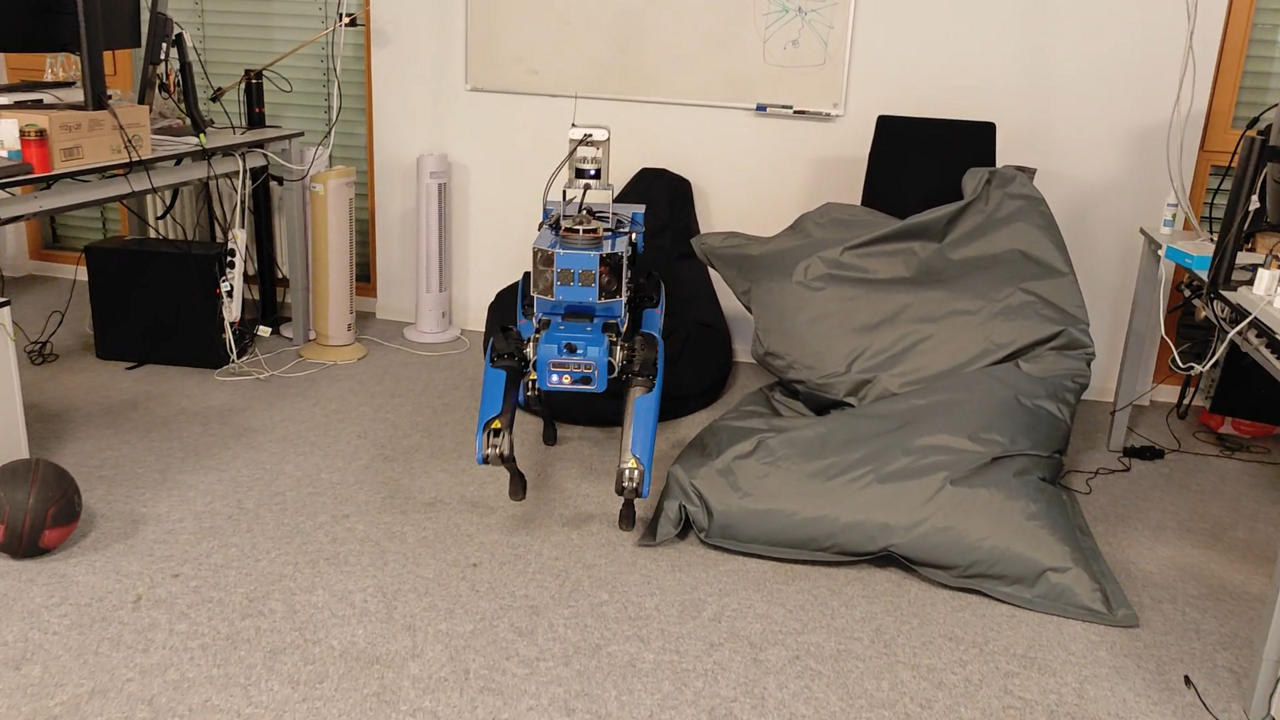}} & 
{\includegraphics[width=\linewidth]{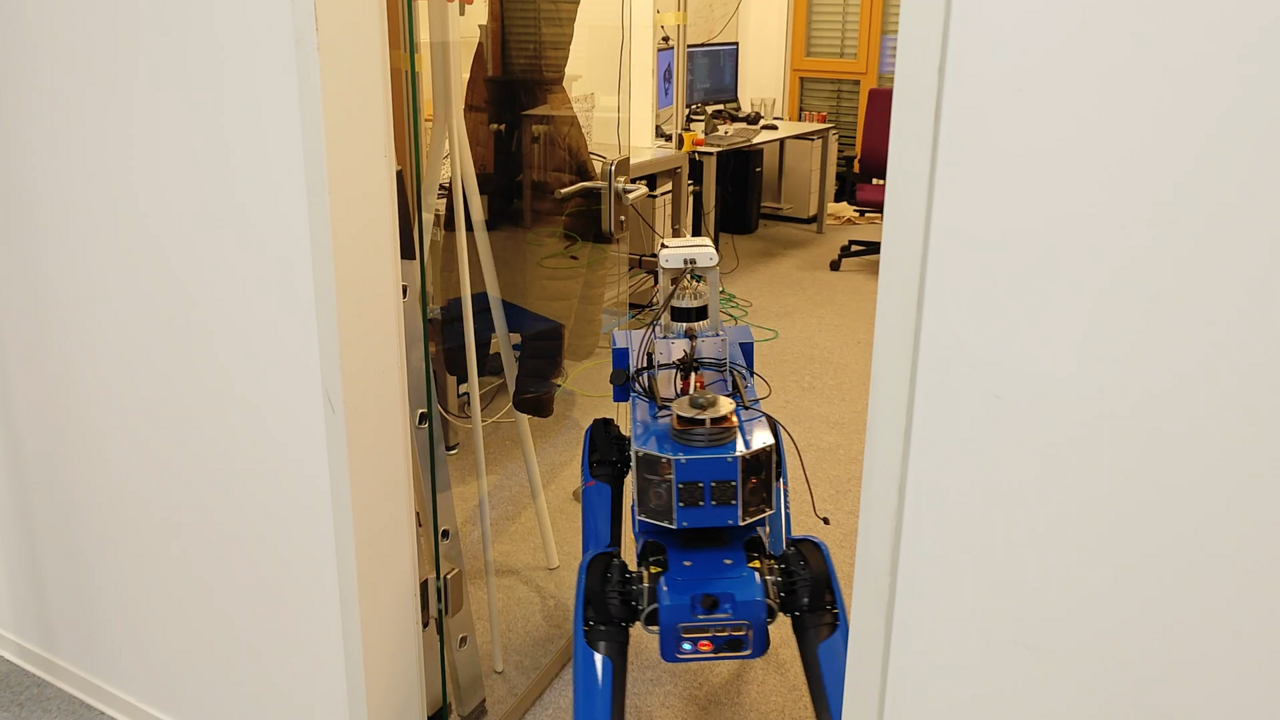}} & 
{\includegraphics[width=\linewidth]{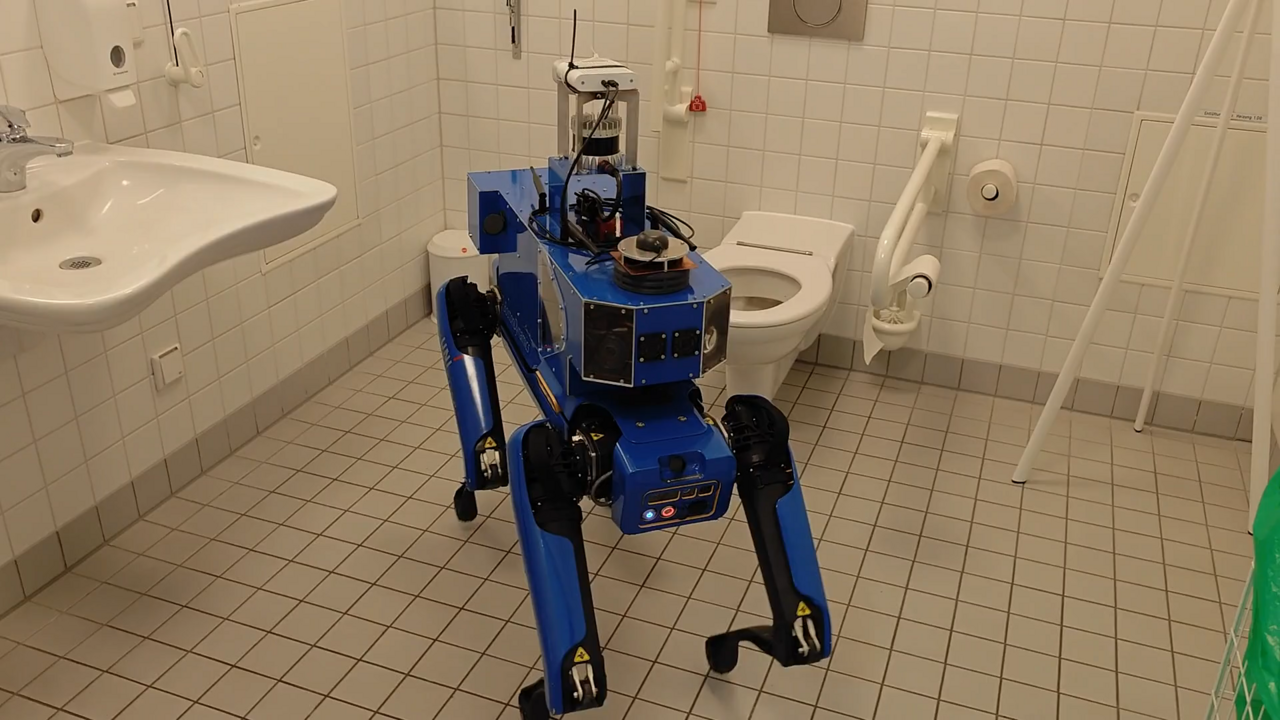}} 
\\
\end{tabular}
}
\caption{Real-World Object Navigation from Language Queries: We show a set of qualitative results of the real-world demonstration trials, which uses a Boston Dynamics Spot to allow for multi-floor traversals. The first row displays the observed scene and the taken path from the start (red) to the goal location (green). The following rows detail the time-wise progression (top-to-bottom).}
\label{fig:real-world-trials}

\end{figure*}

\begin{figure*}
\centering
\footnotesize
\includegraphics[width=0.9\textwidth]{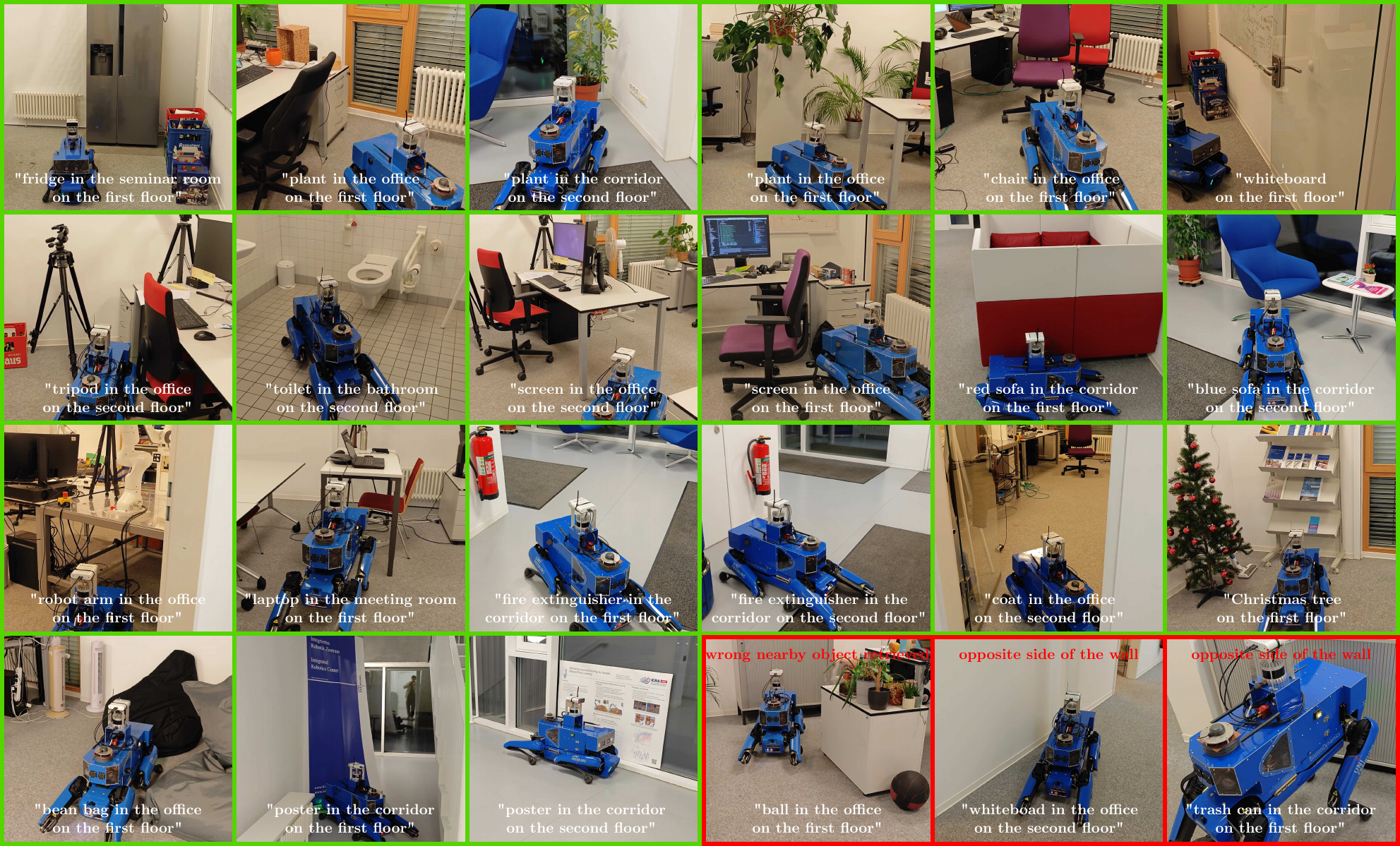}
\caption{\rebuttal{Qualitative visualization of the real-world language-grounded navigation results. Examples masked green denote successful trials, and red masks represent failure cases.}} 
\label{fig:real_world_target_object}
\end{figure*}



\end{document}